\documentclass{article}

\usepackage{arxiv}

\usepackage[T1]{fontenc}
\usepackage{hyperref}
\usepackage{url}
\usepackage{amsfonts}
\usepackage{microtype}
\usepackage{graphicx}
\usepackage{natbib}
\usepackage{doi}

\usepackage{placeins}
\usepackage{algorithm}
\usepackage{algorithmic}

\usepackage{amsmath}
\usepackage{amssymb}
\usepackage{mathtools}
\usepackage{amsthm}

\usepackage{graphicx}
\usepackage{subfigure}

\title{When in Doubt, Think Slow: Iterative Reasoning with Latent Imagination}

\date{}

\author{{Martin Benfeghoul}\\
	Huawei Technologies R\&D\\
        London, United Kingdom\\
	\texttt{martinbenfeghoul@gmail.com} \\
	\And
	{Umais Zahid} \\
	Huawei Technologies R\&D\\
        London, United Kingdom\\
	\texttt{umaiszahid@outlook.com} \\
	\AND
	Qinghai Guo \\
	Huawei Technologies R\&D\\
        Shenzhen, China\\
	\texttt{guoqinghai@huawei.com} \\
	\And
	Zafeirios Fountas \\
	Huawei Technologies R\&D\\
        London, United Kingdom\\
	\texttt{zafeirios.fountas@huawei.com} \\
}

\renewcommand{\headeright}{}
\renewcommand{\undertitle}{}
\renewcommand{\shorttitle}{Iterative Reasoning with Latent Imagination}

\hypersetup{
pdftitle={When in Doubt, Think Slow: Iterative Reasoning with Latent Imagination},
pdfsubject={cs.LG},
pdfauthor={Martin Benfeghoul, Umais Zahid, Qinghai Guo, Zafeirios Fountas},
pdfkeywords={Reinforcement Learning, Iterative Inference, System 2 Reasoning, Perception, Dreamer, Latent Imagination},
}

\begin{document}
\maketitle

\begin{abstract}
In an unfamiliar setting, a model-based reinforcement learning agent can be limited by the accuracy of its world model. In this work, we present a novel, training-free approach to improving the performance of such agents separately from planning and learning. We do so by applying iterative inference at decision-time, to fine-tune the inferred agent states based on the coherence of future state representations. Our approach achieves a consistent improvement in both reconstruction accuracy and task performance when applied to visual 3D navigation tasks. We go on to show that considering more future states further improves the performance of the agent in partially-observable environments, but not in a fully-observable one. Finally, we demonstrate that agents with less training pre-evaluation benefit most from our approach. 
\end{abstract}


\section{Introduction}\label{sec_intro}
Imagine being presented with two images: one depicting a familiar face, the other an unfamiliar object. While the former is instantly recognised, the latter demands deeper scrutiny. This reflects the dual nature of human cognition as described in dual process theory \citep{evans1984heuristic,Kahneman11}. \textit{System 1} encompasses our intuitive, automatic responses, swiftly processing familiar information. In contrast, \textit{System 2} involves a slower, more deliberate form of reasoning, activated when we encounter new or complex situations with higher uncertainty. The latter requires more thoughtful analysis, leveraging abstractions from past experience.

Machine learning algorithms, and neural networks in particular, excel at mimicking \textit{System 1} thinking by efficiently utilising correlations from extensive training datasets for rapid, intuitive inference. Yet, the field faces challenges in developing algorithms capable of \textit{System 2}-like reasoning: a more robust, flexible methodology that supports multi-step, deliberate problem-solving. Such an approach could enable algorithms to learn and adapt to new tasks with greater efficiency, significantly reducing reliance on large training sets \citep{bengio2021system2,du2022IRlearning}. 

A substantial number of approaches which mirror aspects of system 2-like cognition have already been proposed, including planning (e.g. \citealp{garcia1989MPC,coulom2006mcts}), reasoning \citep{du2022IRlearning,wei2022chainOfThought} and, more generally, iterative refinement at inference
\citep{graves2016adaptive,dehghani2018universal,banino2021pondernet,nye2021improving}.
In addition, certain model-based reinforcement learning (RL) methods are also thought to be linked to slow, goal-directed behaviours observed in animals \citep{Sutton1998}. Such methods utilise a learned world model for the planning and learning of optimal actions via imagined trajectories \citep{moerland2022modelbased}. However, their effectiveness hinges on the accuracy of the world model in representing unfamiliar environments, limiting performance without additional training.

In this work, we introduce a novel method that allows for an interplay between system 1 and system 2 perceptual inference in model-based RL agents. Our method enhances the posterior representations of world states, thereby improving task performance independently of traditional planning or learning processes. Drawing from prior research on iterative optimisation in amortised variational inference \citep{marino2017iterativeInference,tschantz2023hybrid}, we introduce an intrinsic inference objective to refine the agent's performance without additional training or introducing new parameterised modules. 
This objective relies on the agent's learnt world model, to compute a Monte Carlo estimation of the expected future observations, using latent imagination. By applying backpropagation to these imaginary rollouts, our method enables the agent to effectively reason about coherence with future states, accessing information gathered in other episodes and practically distilling experience that has been amortised within the world model.

We tested this framework on the recent DreamerV3 agent \citep{hafner2023dreamerv3}, observing a significant performance boost in partially-observable environments, where information is more sparse across states, as well as reducing the well-known amortisation gap in all evaluated scenarios.

Our resulting contributions are as follows:
\begin{itemize}
    \item A novel training-free approach to improve a model-based RL agent with iterative inference, acting at decision-time with latent imagination.
    \item Thorough testing and corresponding analysis of the resulting inference module with various theoretically-justified objectives in visual RL tasks.
    \item A comparison of the impacts of our approach depending on the baseline performance of the agent pre-evaluation.
\end{itemize}

\section{Related Work}\label{sec_background}
\subsection{Amortised Variational Inference}
Variational inference \citep{jordan1998introduction} facilitates the learning of latent variable models by reformulating inference as an optimisation problem \citep{neal1998view,hoffman2013stochastic} with the use of an approximate posterior. It is based around minimising the KL divergence between approximate $q(z|x)$ and true posteriors $p(z|x)$, for observations $x\in \mathcal{D}$ and latent states $z$. To avoid computing the intractable $p(z|x)$, this quantity can be decomposed as:
\begin{equation}\label{eq:VI_KL}
    D_{\text{KL}}\big[q(z|x)||p(z|x)\big] = \text{log}p_{\theta}(x) - \mathcal{L}_{\text{ELBO}}
\end{equation}
with the Evidence Lower BOund (ELBO) $\mathcal{L}_{\text{ELBO}}$:
\begin{equation}\label{eq_ELBO}
    \mathcal{L}_{ELBO} = \mathbb{E}_{q(z|x)}[\text{log} p_{\theta}(x|z)] - D_{\text{KL}}\big[q(z|x)||p_{\theta}(z)\big]
\end{equation}
As the model evidence is independent of the variational distribution, maximising the ELBO with respect to it therefore also minimises the LHS of Eq.~\ref{eq:VI_KL}. Finally, as the latter is also non-negative, we obtain the lower bound on model evidence $\text{log}p_{\theta}(x) \geq \mathcal{L}_{ELBO}$. This property is leveraged by variational autoencoders (VAEs: \citealp{kingma2013autoencoding,rezende2014vae}), which parameterise both the approximation $q_\phi(z|x)$ and the likelihood $p_\theta(x|z)$ and optimise them using Eq.~\ref{eq_ELBO}. Amortisation is applied by employing a shared, learnable function to compute the approximate posterior for each observation, effectively distributing the computational cost across the entire dataset.

\subsection{Connecting Fast and Slow with Iterative Refinement}\label{sec_iter_refine}
Amortised inference models suffer from the \textit{amortisation gap} \citep{cremer2018inference,krishnan2018challenges}, due to the often limited capacity of the neural network used for encoding.
To combat this issue, some hybrid optimisation methods combining amortised (fast) and iterative (slow) inference have been proposed. Such methods generally seek to iteratively refine individual approximate posteriors with respect to Eq.~\ref{eq_ELBO}, or a similar lower bound, using the inference network estimate as a starting point. Examples include iterative refinement \citep{hjelm2018iterRefinement}, iterative optimisation \citep{krishnan2018challenges}, semi-amortised VAEs \citep{kim2018semiamortised}, models for control \citep{millidge2020reinforcement,tschantz2020control,marino2021iterative,millidge2020relationship}, perception \citep{marino2018iterativeAmortised}
 and modern approaches to predictive coding \citep{tschantz2023hybrid,zahid2023sample}. 
While the inference objectives and methods used vary slightly across these approaches, the key idea remains the same: iteratively refine approximate posterior estimates with objectives based on variational inference, using an amortised inference network to provide the starting point. 

Iterative Amortised Inference models, as introduced by \citet{marino2017iterativeInference,marino2018iterativeAmortised}, go one step further and seek to \textit{learn} how to iteratively update approximate posterior estimates by encoding the corresponding gradients. This class of models, along with the other methods discussed above, manage to achieve approximate posterior estimates that are closer to the true posterior.

Finally, this interplay between fast and slow optimisation of inference has also been conceptualised outside the realm of variational inference. It was originally proposed for planning in RL \citep{sutton1991dyna,schmidhuber1990making}, with other notable mentions including \citet{silver2016mastering} and \citet{anthony2017thinking} while, more recently, it was also introduced in large language models \citep{lin2023swiftsage,trinh2024solving}.

\subsection{Iterative Reasoning}
A slightly different family of algorithms which resemble system 2-like processing involve reasoning, realised via recurrent computation with deep neural networks \citep{graves2016adaptive,bolukbasi2017adaptive,chung2017hierarchical,eyzaguirre2020differentiable,veerabadran2023adaptive,schwarzschild2021learnAdaptive,dehghani2018universal}. Central to these algorithms is the integration of meta-learning, typically through an additional module that determines computational steps or network component activation based on input. However, these methods have notable limitations. They are prone to instability and sensitivity to hyper-parameters \citep{banino2021pondernet}, and require specific model architectures and additional inference modules. These are all requirements that can significantly restrict the methods' generalisability.

Closest to our work, \citet{du2022IRlearning} implements iterative reasoning as energy minimisation (IREM), by training a neural energy-based model to parameterise an optimisation landscape over all provided data, then iteratively applying gradient steps to find the minimal energy solution for a given input. This method showed promising results when applied to algorithmic reasoning tasks, illustrating that reasoning can be viewed as a set of optimisation steps in a continuous energy landscape. This is a crucial insight for our current work, as we propose that optimising an expected variational lower bound, via performing latent imaginary transitions, can be also viewed as a form of iterative reasoning.

\section{Method}\label{sec_method}
\subsection{DreamerV3}\label{sec_dreamerv3}

DreamerV3 \citep{hafner2023dreamerv3} is a model-based RL agent which learns a world model through direct interaction with its environment, then uses this model to learn an actor-critic entirely in model-generated, "imagined" trajectories. 

The world model is made up of six different modules, each parameterised by a neural network. Its main goal is to learn a compact representation, latent state $z_t$, of its sensory inputs $x_t$ using an auto-encoder architecture \citep{kingma2013autoencoding,rezende2014vae}. A deterministic sequence model with recurrent state $h_t$ is then used alongside a dynamics predictor to predict the sequence of these stochastic representations, given past actions $a_{t-1}$. Therefore, the dynamics predictor provides a prior latent state distribution for $\hat{z}_t$, while the encoder outputs the posterior distribution $z_t$. The concatenation of $h_t$ and $z_t$ is referred to as the model state, which is then used to predict rewards $r_t$, episode continuation flags $c_t \in \{0, 1\}$, and decoder-reconstructed observations $\hat{x}_t$. This gives the following structure, with $\phi$ denoting the collective parameters that make up this world model:
{\allowdisplaybreaks
\begin{subequations}
\begin{align}
    &\text{Sequence model:} && h_t = f_{\phi}(h_{t-1}, z_{t-1}, a_{t-1})\label{eq_seqModel}\\
    &\text{Encoder:} && z_t \sim q_{\phi}(z_t | h_t, x_t)\label{eq_enc}\\
    &\text{Dynamics predictor:} && \hat{z}_t \sim p_{\phi}(\hat{z}_t | h_t)\label{eq_dyn}\\
    &\text{Reward predictor:} && \hat{r}_t \sim p_{\phi}(\hat{r}_t | h_t, z_t)\label{eq_reward}\\
    &\text{Continue predictor:} && \hat{c}_t \sim p_{\phi}(\hat{c}_t | h_t, z_t)\label{eq_cont}\\
    &\text{Decoder:} && \hat{x}_t \sim p_{\phi}(\hat{x}_t | h_t, z_t)\label{eq_dec}
\end{align} 
\end{subequations}
}

\subsection{Iterative Inference (II)}
Our approach to II seeks to iteratively improve the agent's representation of its current state with respect to its predicted future, while avoiding further training and parameterised modules. In order to do so, we have chosen to take local gradient steps with respect to an \textit{inference objective} $\mathcal{L}_{Obj}$. We iteratively apply these updates directly to the agent's initial representation of its state, rather than amortising updates to the corresponding model parameters. This allows us to make local improvements specific to the current state without changing the overall world model, and follows local updates observed in iterative refinement methods (\ref{sec_iter_refine}) to reduce the amortisation gap. 

Common targets for iterative optimisation include the KL divergence between approximate and true (model) posteriors. To render this tractable, II generally appeals to (non-amortised) optimisation of an ELBO with respect to the approximate posterior's distributional parameters. Here, we instead consider the impact of optimising surrogate objectives inspired by the \textit{future} expected model evidence, with the goal of optimising states towards something that is consistent with the projections expected by the agent's world model. In essence, we hypothesise that posterior state estimation improves when the imagined rollouts these states entail are not unexpected by the agent.

The expected model evidence for future states beyond the current time-step $t$, and up to episode termination $T$, may be decomposed as followed, under an expectation over future states:
\begin{subequations}\label{eq_future_model_evidence}
\begin{align}
\mathbb{E}&_{p(x_{t:T},z_{t:T}|z_t)} \mathcal{L}_{ELBO}=\notag\\
&~~\mathbb{E}_{p(x_{t:T},z_{t+1:T}|z_t)}\mathbb{E}_{q(z|x)q(\theta|x)} \log p(x|z,\theta) \label{eq:2a}\\
&~~-\mathbb{E}_{p(x_{t:T},z_{t+1:T}|z_t)}\mathbb{E}_{q(\theta|x)}D_{KL}\big[q(z|x)||p(z|\theta)\big] \label{eq:2b}\\
&~~-\mathbb{E}_{p(x_{t:T},z_{t+1:T}|z_t)}D_{KL}\big[q(\theta|x)||p(\theta)\big] \label{eq:2c}
\end{align}
\end{subequations}

Given the lack of access to future observations, we adopt surrogate objectives, which we outline in section \ref{sec_inferenceObjectives}. For each iterative gradient step, the world model is used to "roll-out" a forward (future) trajectory of imagined latent steps of length $\lambda$, starting from our current state representation. Our inference objective is computed over this entire trajectory. We do so using trajectory sampling \citep{BARTO199581} given the agent's learned greedy policy. As these trajectories, which we will refer to in this work as \textit{rollouts}, are stochastic, we take the average value of the inference objective over a small number $s$ rollouts. This represents a low-sample Monte-Carlo estimate of the expectation over all possible rollouts, given our current state and policy. 

Finally, due to DreamerV3's stochastic latent states and actions, we expect it may be difficult to compare the results of our inference module with that of the default agent. To combat this issue and reduce noise within and between episodes, we move to a more deterministic selection of next states and actions. We take $z_t$ to be the mode of the latent state distribution before applying II, as well as after every update. After the II updates, we do the same thing for the Actor's output action distribution before returning both the updated $z_t$ and $a_t$ to the default agent for the next environment step. This is also applied when evaluating the default agent to ensure a fair comparison with this baseline.
 
\begin{figure}[ht]
    \centering
    \begin{minipage}{.7\linewidth}
        \begin{algorithm}[H]
        \caption{Iterative Inference in DreamerV3}\label{alg_II}
        \begin{algorithmic}
            \STATE \textbf{Input:} Observation $x_t$ received at time $t$.
            \STATE \textbf{Input:} Recurrent state $h_t$.
            \STATE \textbf{Output:} Updated $h_t$,$z_t$,$a_t$
                \STATE $z^0_t = \text{argmax}_{z}\ q_{\phi}(z^0_t = z | h^0_t, x_t)$
                \FOR{iteration $i \in \{0,1,...,n-1\}$}
                    \FOR{$s$ sampled trajectories}
                        \FOR{rollout state $j \in \{1,2,...,\lambda\}$}
                            \STATE $h^i_{t+j} \gets f_{\phi}(h^i_{t+j-1}, \hat{z}_{t+j-1}, a_{t+j-1})$
                            \STATE $\hat{z}_{t+j} \sim p_{\phi}(\hat{z}_{t+j} | h^i_{t+j})$
                            \STATE $a_{t+j} \sim \text{Actor}(a_{t+j} | h^i_{t+j}, \hat{z}_{t+j})$
                        \ENDFOR
                        \STATE $\mathcal{L} \mathrel{+}= \mathcal{L}_{Obj}(\hat{z}_{t:t+\lambda})/s$
                    \ENDFOR
                    \STATE $\mathcal{L} \mathrel{+}= \mathcal{L}_{Reg}(z^0_t,z^i_{t})$
                    \STATE $h^{i+1}_t \gets \alpha \cdot \Delta_{\phi} \mathcal{L}$
                    \STATE $z^{i+1}_t = \text{argmax}_{z}\ q_{\phi}(z^{i+1}_t = z | h^{i+1}_t, x_t)$
                \ENDFOR
                \STATE $a_{t} = \text{argmax}_{a}\ \text{Actor}(a_t = a | h_{t+i}, \hat{z}_{t+i})$
                \STATE \textbf{return} $h^n_t$, $z^n_t$, $a_t$ 
        \end{algorithmic}
        \end{algorithm}
    \end{minipage}
\end{figure}

Our application of II to the DreamerV3 agent is best summarised in Algorithm \ref{alg_II}. Updates are applied directly to $h_t$ and are scaled by a \textit{step size} $\alpha$. We could also simply update the action distribution in order to pick actions which lead to states that minimise our objective. However, by updating the underlying model state used for action selection, we leave room for improving state representations in two ways. First, by reducing a potential amortisation gap in our current state representation, which will be measured by our reconstruction metrics. Secondly, by indirectly influencing the resulting action selection in a way which may lead to better trajectories, influencing overall task performance as well as reconstruction metrics of future states. Note that we backpropagate gradients through the actor across rollouts to facilitate the latter.

\subsection{Inference Objectives}\label{sec_inferenceObjectives}
We look to maximise model evidence via its lower bound in Eq.~\ref{eq_future_model_evidence}. The decomposed terms in this equation can be contextually interpreted within the framework of information theory, drawing parallels with similar concepts in intrinsic reward in RL \citep{fountas2020deepActive,rhinehart2021information}: Eq.~\ref{eq:2a} maximises the likelihood of observations $x$, while \ref{eq:2b} and \ref{eq:2c} seek to minimise state and parameter uncertainty (respectively) in the corresponding states. As the true observations $x$ for future, and therefore unseen, states are unavailable to us at decision-time, we will focus on approximating and minimising state and parameter uncertainty. Furthermore, we apply each term as a separate inference objective, in order to further analyse the impact of each one on our agent.

Information gain (IG) \citep{lindley1956information,mackay1992informationObj} represents the amount of new information that we can expect to acquire by taking a measurement compared to our prior knowledge for the corresponding data point. This can be maximised to select the data points which will improve our algorithm the most when learnt, minimising future uncertainty on similar data points. This is seen in active learning \citep{burr2010activelearning}, active inference \citep{FRISTON20168activeinfandlearning}, and notions of intrinsic motivation in RL \citep{Schmidhuber1991curiosity,bellemare2016intrinsic,aubret2019surveyIntrinsic}. As we are focused on maximising decision-time performance rather than learning and exploration, we will seek to minimise this quantity, as is the case when maximising model evidence (Eq. \ref{eq_future_model_evidence}). 

\textbf{State Information Gain (SIG)}.
We therefore implement an approximation to eq.~\ref{eq:2b}, \textit{state information gain}, using our model's prior and posterior latent states.
As future observations $x$ are needed to calculate the approximate posterior $q_{\phi}(z_t | h_t, x_t)$ with the encoder, we make a further approximation by instead using the decoder-reconstructed prior. We then take the Kullback-Leibler divergence between these two terms, at each II iteration $i$:
\begin{equation}\label{eq_IG}
    \mathcal{L}_{SIG} = \frac{1}{\lambda} \sum_{j=0}^{\lambda} KL(q_{\phi}(z^i_{t+j} | h^i_{t+j}, \hat{x}_{t+j}) || p_{\phi}(\hat{z}_{t+j} | h^i_{t+j}))
\end{equation}

\textbf{Parameter Information Gain (PIG)}.
Latent disagreement, as implemented in the model-based Plan2Explore agent \citep{sekar2020plan2expl} for efficient exploration, provides an approximation to eq.~\ref{eq:2c}, \textit{parameter information gain}. It is calculated by training a bootstrap ensemble \citep{Breiman1996bagging} of $K$ predictive models with parameters $\{w_k | k \in [1;K]\}$ which predict the latent state $z_t$ given the recurrent state $h_t$. Each model has a different initialisation and is trained on observations in a different order from the others. We can therefore use the variance in the ensemble predictions to measure the parameter IG of a trajectory:
\begin{equation}\label{eq_LD}
    \mathcal{L}_{PIG} = \frac{1}{\lambda} \sum_{j=0}^{\lambda} Var(\{\mathbb{E}[w_k,h^i_{t+j}] | k \in [1;K]\})
\end{equation}

\textbf{Entropy (ENT)}.
Finally, stepping away from the ELBO and looking at uncertainty in general, a well-known and easy way to measure the uncertainty in predictions is to look at the entropy of the output distribution \citep{CEShannon}. While maximum entropy exploration \citep{hazan2019provably} has been used to drive efficient exploration in an RL environment, this quantity does not directly relate to iterative inference and our model evidence. We include it for comparison as a widely-known and accessible measure of uncertainty, as we have linked our other objectives to such measures:
\begin{equation}\label{eq_entropyObj}
    \mathcal{L}_{ENT} = \frac{1}{\lambda} \sum_{j=0}^{\lambda} \mathbb{E}[-\text{log}\ p_{\phi}(\hat{z}_{t+j} | h^i_{t+j}))]
\end{equation}

\subsection{Regularisation}
Without the likelihood term (Eq. \ref{eq:2a}) in our objectives, we need to ensure that our updated states are still representative of the current observed state.  We therefore introduce a cheap KL loss term between the initial latent state estimate and the updated latent state at iteration $i$, to act as a regulariser. We add free bits to leave room for some non-penalised changes to the representation:
\begin{equation}\label{eq_klReg}
\mathcal{L}_{Reg} = max(1, D_{KL}[q_{\phi}(z^{0}_t | h^{0}_t, x_t) ||q_{\phi}(z^{i}_t | h^{i}_t, x_t)])
\end{equation}

\subsection{Environments and Tasks}
In order to study our framework's behaviour when presented with various tasks, we chose to run our experiments over three different visual environments: two DeepMind Lab (DMLab) \citep{beattie2016deepmind} tasks, and one Atari \citep{bellemare2013atari} task. We also add our own task, built using Miniworld \citep{gym_miniworld}, for a more granular analysis of agent behaviour. Both DMLab and Miniworld are first-person, 3-Dimensional (3D) environments. Accurately navigating such a visually and spatially complex, partially-observable (PO) environment should require a good grasp of the environment dynamics, as well as some form of short-term memory in order for state representations to take into account objects and entities outside of the field of view. Atari games on the other hand, are 2-Dimensional and  provide a full view of the environment at all times, hence it is fully-observable (FO) and can therefore rely much more on the encoder alone to provide more complete and accurate state representations. Furthermore, observations are often a mostly static image with the exception of a few small moving objects or characters. Testing our approach on the latter is expected to provide a valuable performance comparison with the more complex 3D tasks.  

More specifically we chose DMLab's "Collect Good Objects" (DMLab 1) task as the first 3D task to provide a benchmark on a relatively visually and spatially simpler task. The second DMLab task, "NatLab Fixed Large Map" (DMLab 2) is a much more visually and spatially complex task chosen to further challenge our agent in those domains. For the last 3D task, we designed a visually simple MiniWorld task, but which requires a much finer navigation of its environment in order to observe high rewards and avoid negative ones. Finally, we chose "Alien" for the 2D Atari task, in order to assess our method in a FO environment.

\section{Results}\label{sec_results}
We evaluate and compare our approach to the default DreamerV3 agent at multiple stages during its training. We refer to the "pre-training volume" as the amount of environment steps the agent has undergone pre-evaluation. Our results are measured in two ways: overall performance and immediate impact. Overall performance looks at the "long-term" effect of our approach on entire episodes. On the other hand, immediate impact metrics only measure the "short-term" difference before and after applying II to the agent's initial state representation for a single observation. 
In order to measure the accuracy of state representations with respect to the true observations, we mainly employ two reconstruction metrics: the Peak-Signal-to-Noise-Ratio (PSNR) and the Structured Similarity Index (SSIM) \citep{ssim2004}. These will measure various differences between the true observed $x_t$ and the decoder-reconstructed $\hat{x}_t$.

\subsection{Overall Performance}

\begin{figure}[bt]
    \centering    \includegraphics[width=0.9\linewidth,height=0.9\textheight,keepaspectratio]{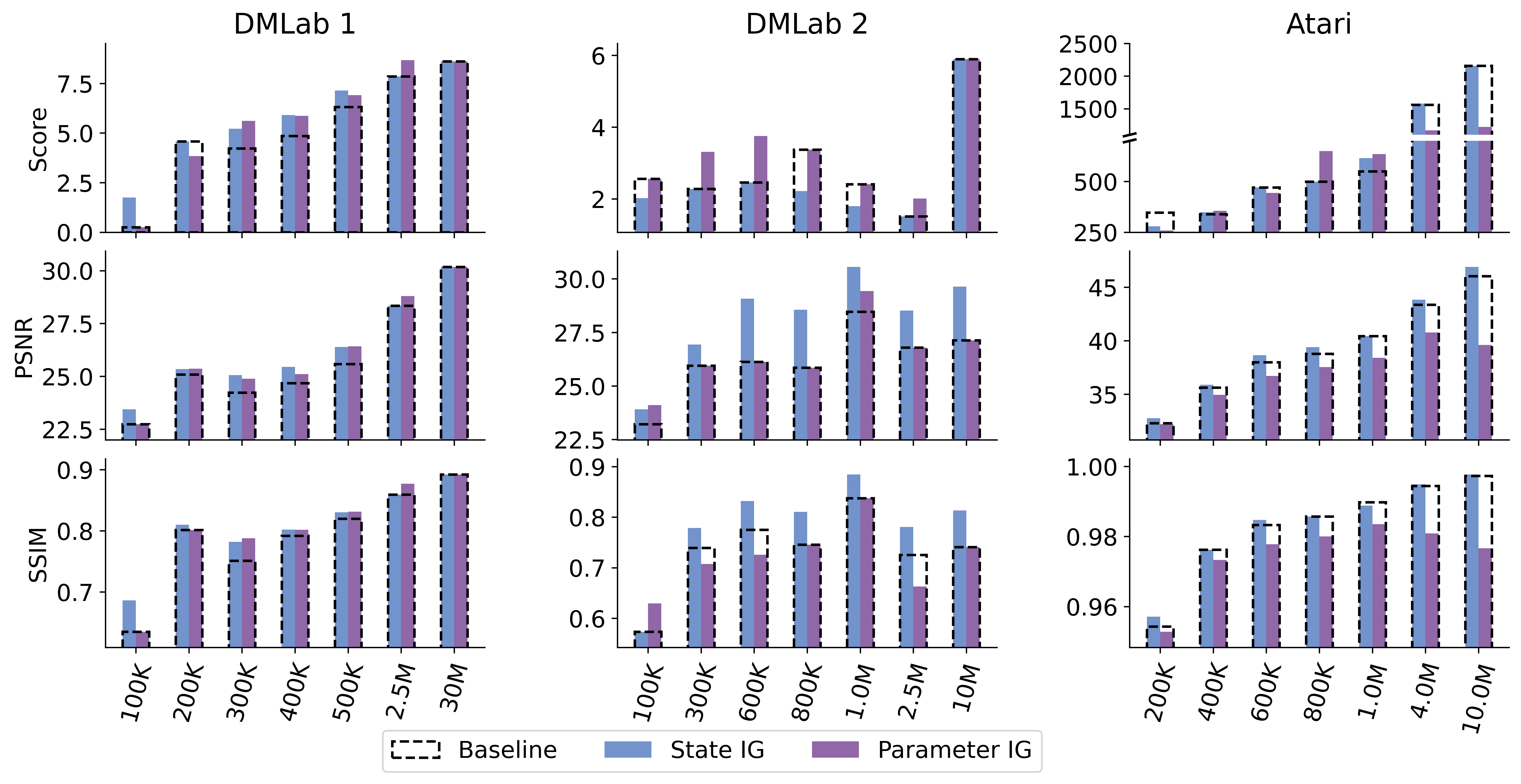}
    \caption[Inference Objective Performance Comparison v. Pre-Training Volume]{Comparison of the average performance metrics for iterative inference with two inference objectives versus the default DreamerV3 agent (Baseline). For each objective, we take the results of the rollout length $\lambda$ with the best, \textit{significant} improvement compared to the baseline (ie. a p-value below 5\%), across all metrics, assuming the same value as the baseline if the result isn't significant. Experiments are run over 100 episodes each.}
    \label{fig_MainComparison}
\end{figure}

Looking at reconstruction metrics over entire episodes (Fig. \ref{fig_MainComparison}) we can see a consistent improvement over the default DreamerV3 agent for both DMLab tasks, as well as in Atari. 

DMLab 1, shows a consistent improvement over baseline across most pre-training volumes with both objectives, and seems to plateau after enough training pre-evaluation. DMLab 2, our most visually and spatially diverse task, shows especially large improvements in these metrics with the State IG objective across all pre-training volumes from 300K steps onwards. The sustained performance at higher training volumes might be attributed to the task's large environment and diverse observations, enabling exploration of mostly unencountered states even at higher volumes. Parameter IG however does not show such consistent improvements, and can even make reconstruction metrics worse in DMLab 2.

Atari also showed very consistent, yet much smaller improvements with State IG across most pre-training volumes, despite the baseline agent's already near-perfect reconstructions. The latter is likely explained by an amortisation gap, which is then reduced by II, as seen in iterative refinement methods. 
Parameter IG, on the other hand, almost always decreased reconstruction accuracy, with an especially large drop in performance in larger volumes. We note that the similarity between observations, along with the FO nature of this environment may be the reason for such limited performance with this objective compared to the other tasks.

Regarding the best performing rollout lengths, II consistently showed its best performance with the longer $\lambda=8,16$ in our DMLab tasks. On the other hand, almost all the biggest improvements over baseline were with $\lambda=1$ in Atari. In fact, larger rollout lengths often showed a \textit{decrease} in performance in this environment.

\subsection{Immediate Impact}

\begin{figure}[tb!]
    \centering
    \includegraphics[width=1.0\linewidth]{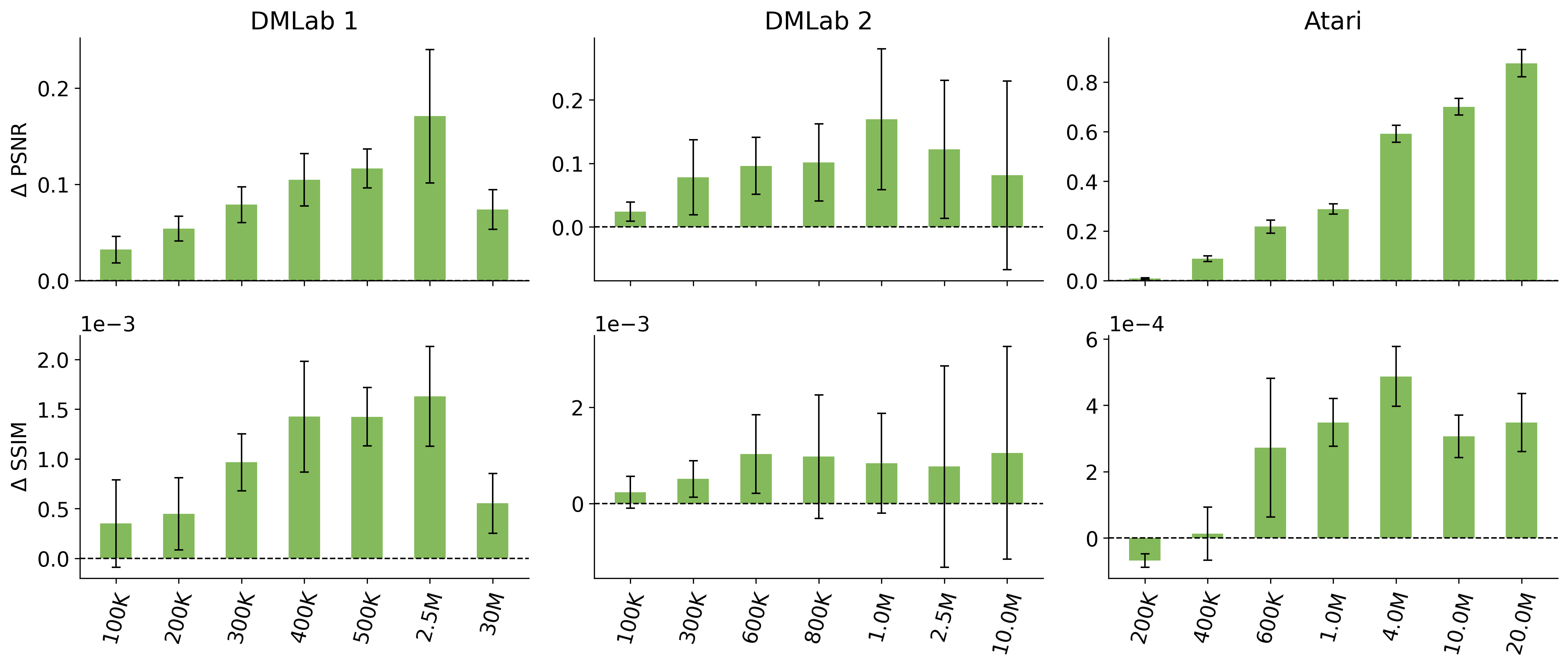}
    \caption{Immediate improvement on the current latent state reconstruction metrics when the State IG objective is used with rollout length $\lambda=1$. Values and standard deviation are measured over all environment steps for 100 episodes each.}
    \label{fig_IG_diff_r1}
\end{figure}

When looking at the immediate impact of II on our reconstruction metrics (Fig. \ref{fig_IG_diff_r1}), we found that State IG was the only objective to achieve significant improvements on the agent's initial representation directly after applying our updates. While we have seen the Parameter IG objective improve such metrics when looking at overall episodes, this objective either matched or decreased the accuracy of initial state reconstructions when applying II for all rollout lengths. 

Furthermore, while multiple rollout lengths achieved significant improvements, the difference observed in the current state consistently decreases as the rollout length increased. In fact, for the Atari task, longer rollout lengths showed a consistent \textit{decrease} in reconstruction metrics. This is likely due to the fact that, for a shorter rollout, the current state will have a much bigger impact on the averaged objective value. This is also consistent with overall episode metrics showing better performance for larger values of $\lambda$, given that longer rollouts take more future states into account.

\subsection{Further Analysis: Rollout Length}

Overall performance metrics show conflicting conditions in achieving the best performance compared to immediate impacts. The fact that immediate impacts are only significant with the State IG objective and greatest with $\lambda = 1$ across all experiments, does not explain the larger improvements observed in overall performance at longer rollout lengths, and with the Parameter IG objective. This suggests that II affects the current state representation differently, depending on the rollout length. 

\begin{figure}
    \centering
    \includegraphics[width=\linewidth]{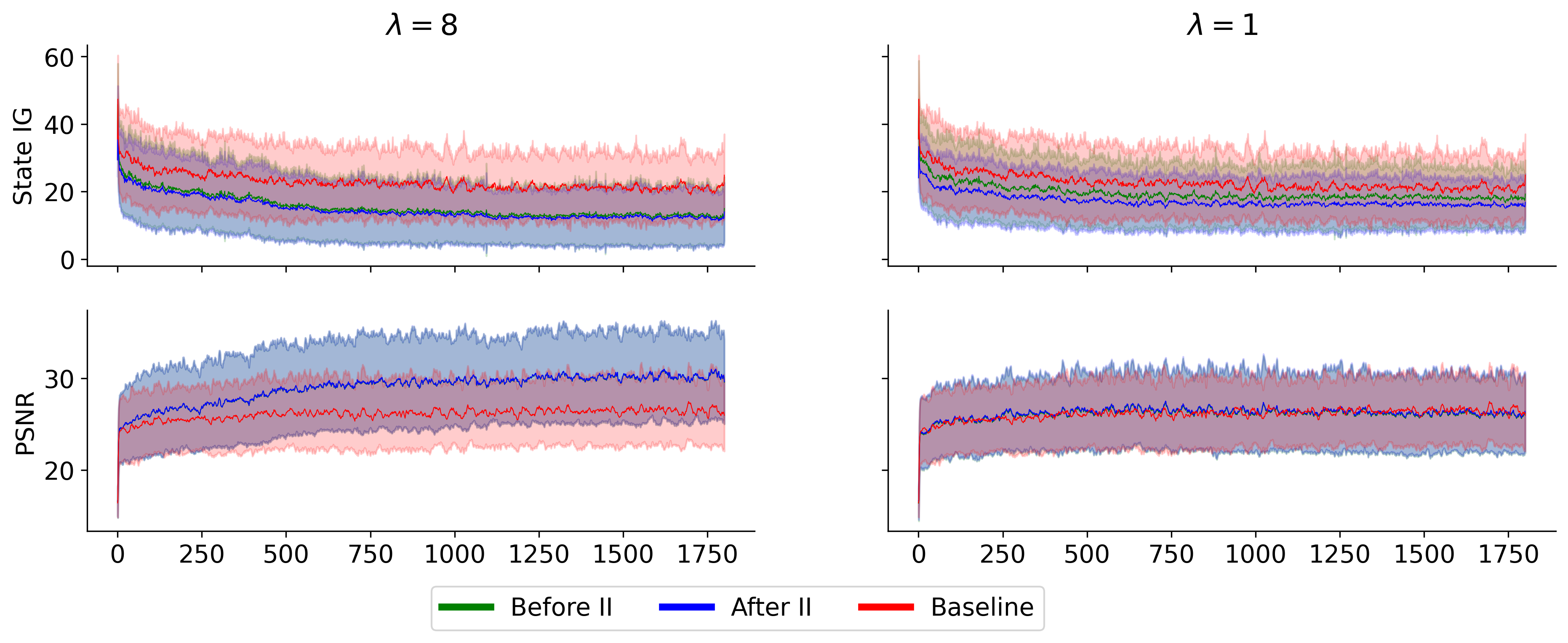}
    \caption{Comparing the different impact on the inference objective and reconstruction metrics over environment steps, with different rollout lengths. All plots show II applied with the State IG objective to DMLab 2 after 600K pre-training steps, compared to the baseline agent. Values are averaged across 100 episodes.}
    \label{fig_natlab_trajectories}
\end{figure}

Together with our previous observations, Figure \ref{fig_natlab_trajectories} further illustrates such differences depending on the rollout length used. We observe that $\lambda = 1$ shows a greater immediate impact on the current objective, but manages a lesser impact on long-term values. On the other hand, $\lambda=8$ shows a greater improvement across environment steps with almost no discernible immediate impact.

We might explain the immediate impacts observed at short rollout lengths with a closing of the amortisation gap, as seen in iterative refinement. On the other hand, the large improvements on overall performance do not seem to be explained simply by an accumulation of such small improvements to the current state reconstructions. Instead, we observe a change of trajectory towards future states with more accurate representations. 

It therefore seems that, given enough information on future states, our updates on the current state are also able to influence the actor's action selection in a way that improves the reconstruction metrics of the states that will follow, rather than just the current one. While a short rollout length manages greater improvements to current state reconstructions, it does not seem to change the agent's behaviour to the same extent. Interestingly, this contrast is clear within our PO environments but less so in the FO Atari task, for which the overall performance and immediate impact behaviours seem closely related.

Of course the two behaviours observed are unlikely to act entirely independently, with smaller rollouts sometimes showing (albeit smaller) overall improvements (eg. in Atari), and vice versa. But the greatest improvements seen from each are still clearly divided by the rollout length. Finally, we note that the more visually and spatially complex PO tasks seem to be the ones benefiting most from longer rollouts, with the FO Atari task showing the least improvements in that regard. Instead, Atari's improvements seem likely due to an accumulation of small immediate impacts.

\subsection{Task Performance}

Both State IG and Parameter IG showed an improvement in DMLab 1 task performance at lower volumes ($\leq 2.5M)$. On the other hand, both DMLab 2 and Atari seem to do best with Parameter IG for this metric, despite its lower performance across reconstruction metrics. Furthermore, Fig. \ref{fig_MainComparison} suggests that task performance is mostly improved in lower performing agents, and improving reconstruction metrics does not always improve task performance.

In order to dig further into the conditions for improved task performance, we make use of our Miniworld task. With full control over this environment, we are able to directly compare baseline and II performance in single episodes by ensuring the exact same initialisation and environment layout. This allows us to further investigate any potential differences in the behaviour of II with respect to the baseline agent.

\begin{table}[htb!]
    \centering
    \caption{Miniworld task performance after 1M pre-training steps, in episodes with the same initialisation. "Threshold" shows the maximum baseline agent score with which we selected episodes. Metrics shown are the proportion of episodes scoring below threshold, average episode scores for baseline, State IG, and Parameter IG, and the best rollout length corresponding to the results shown for SIG and PIG respectively. Bold values are significant with respect to baseline (p-value $< 5\%$).}
    \label{tab_roboNav_2}
    \vspace{0.1in}
    \centerline{
    \begin{tabular}{c c c c c c}
        \hline
        \textbf{Threshold} & \textbf{\% Ep.} & \textbf{Baseline} & \textbf{SIG} & \textbf{PIG} & $\mathbf{\lambda}$ \\
        \hline
        All & 100\% & $0.68$ & $0.66$ & $0.67$ & $16|3$\\
        $< 0.90$ & 67\% & $0.55$ & $\mathbf{0.65}$ & $\mathbf{0.64}$ & $1|3$\\
        $< 0.75$ & 49\% & $0.44$ & $\mathbf{0.64}$ & $\mathbf{0.62}$ & $1|3$\\
        $< 0.50$ & 17\% & $0.19$ & $\mathbf{0.60}$ & $\mathbf{0.57}$ & $1|3$\\
        $< 0.25$ & 8\% & $-0.11$ & $\mathbf{0.61}$ & $\mathbf{0.62}$ & $1|8$\\
        $< 0.00$ & 4\% & $-0.20$ & $\mathbf{0.61}$ & $\mathbf{0.62}$ & $1|8$\\
        \hline
    \end{tabular}
    }
\end{table}

Table \ref{tab_roboNav_2} shows significant improvements in episode scores with both objectives. The biggest improvements are observed in episodes where the baseline agent does worst. On the other hand, the opposite is true when looking at episodes where the baseline agent does best, with II often doing worse. 

In fact, all rollout lengths showed significant improvements on all metrics in episodes with lower baseline performance. State IG seemed to achieve its highest scores with $\lambda=1$ in such episodes, with $\lambda=16$ a close second. On the other hand, $\lambda=16$ outperformed shorter rollouts when looking at episodes where the baseline agent scores higher than the threshold. Parameter IG consistently scored highest with $\lambda=3,8$ in both settings. Finally, reconstruction metrics consistently improved baseline metrics in all episodes, achieving similar values regardless of baseline performance.

Overall, results show that II is able to maintain scores close to the overall average baseline performance, even in episodes where the baseline agent may act very poorly. This observation is consistent across all pre-training volumes which we evaluated (400K-10M). This further shows that our approach adjusts the agent's behaviour to act in a more familiar way. This is an exciting result in the case where we want to consistently exploit an agent's previous experience when faced with a less familiar setting.

\subsection{Entropy as an Inference Objective}

The entropy objective (eq. \ref{eq_entropyObj}) did not yield significant improvements to any of our performance metrics across all pre-training volumes (100K-30M) for the DMLab "Collect Good Objects" task. In fact, some results even suggest an increase in variance and a decrease in performance for most metrics and rollout lengths on the agents with larger pre-training volumes, although not statistically significant. This further justifies our focus on maximising model evidence.

\section{Discussion}\label{sec_discussion}
The results discussed in the previous sections show that we are able to consistently improve the baseline agent in visual tasks with at least one of our objectives. This is particularly notable given DreamerV3's current dominance in visual RL benchmarks. Although our method's efficacy is somewhat reliant on DreamerV3's robust world model, its potential applicability extends beyond this specific agent to a broader range of model-based RL algorithms. Hence, our results encourage further exploration of this approach as a way to enhance system 2-like reasoning in various computational models.

Furthermore, our results indicate that enough further training of the baseline agent may beat our approach in task performance. However, II can still show improvements to reconstruction metrics even at higher pre-training volumes. Both of these observations make it a promising approach in a setting where new data is very expensive or simply unavailable. A fitting example application for such results may be in a physical robot seeking to navigate less familiar environments.

\subsection{Future Work}

Motivated by the maximisation of future model evidence, here we explored distinct terms of Eq.~\ref{eq_future_model_evidence} as inference objectives. Although the likelihood term (Eq.~\ref{eq:2a}) was omitted due to its dependency on real future observations, a hybrid approach could incorporate this term only for the current step $t$, where the observation $x_t$ is known. Setting $\lambda=0$ and considering all terms of Eq.~\ref{eq_future_model_evidence} brings our method in line with predictive coding \citep{rao1999predictive}. Hence, through this lens, our method can be viewed as an extension of the approach by \citet{tschantz2023hybrid}, combining the maximisation of both current and future expected model evidence. Finally, another relevant and exciting extension to this work concerns making use of the critic's predicted reward signal to directly integrate the inference of actions, as a form of planning.

A second direction for future work concerns improving the ability of the agent's world model to predict future states that contain useful information for reasoning. This can be done by modelling state transition dynamics based on the sequence of highly surprising events, as opposed to transitioning at every time step of the environment's clock. \citet{zakharov2021episodic} termed this method subjective-timescale models and applied it to an early version of Dreamer \citep{hafner2019dreamer}, to facilitate latent imagination and improve credit assignment. Having the ability to perform rollouts by temporally jumping to the next point of a significant environment change can minimise the effect of accumulated noise and substantially reduce the computational requirements of our method \citep{zakharov2022variational}.

Taking further inspiration from Iterative Amortised Inference, one could also look for a method for amortising the iterative process presented here. In our method, we avoid amortisation to the world model parameters in order to keep updates entirely local. However, we could introduce a similar \textit{optimiser} network \citep{marino2018iterativeAmortised} which takes in the inference objective's gradients and returns the updates which are to be applied to our current state representation. Note that this approach would require further training for the resulting framework to work best, which is not in keeping with this work's focus on a training-free approach.

Finally, we expect that our approach will be most effective in settings where the baseline agent performs worst. Whilst our results have confirmed this, our experiments do not discern states where the agent is \textit{in doubt} from more certain ones and II is therefore applied at every environment step. Introducing a mechanism to dynamically identify such states at decision time, adjust the rollout depth $\lambda$, and halt when a threshold reaches the local minima of the optimisation landscape \citep{du2022IRlearning,tschantz2023hybrid}, would allow us to save computation and potentially further improve overall performance by only applying II when necessary.

\section{Conclusion}\label{sec_conclusion}
In conclusion, our approach to II applied to states with latent imagination demonstrates a consistent improvement on the default DreamerV3 agent across multiple visual RL tasks, with no further training. With the right objective and rollout length, this improvement holds across all metrics, on a variety of pre-training volumes. We also demonstrated that taking more future states into account leads to a more significant change in trajectory and a larger impact on overall performance. Our findings further show that this does not seem to hold in fully-observable environments. Finally, we observed that an agent benefits most from II in unfamiliar settings. We therefore consider our work to be promising in settings where new data is scarce or expensive.

\section*{Impact Statement}
This paper presents work whose goal is to advance the field of Machine Learning. There are many potential societal consequences of our work, none of which we feel must be specifically highlighted here.

\section*{Acknowledgements}
The authors would like to thank Andre Vaillant for his support during preliminary testing of the Atari environment.

\newpage
\appendix

\section{Experimental Settings}\label{app_experimental_settings}
\subsection{Pre-Training}
As previously mentioned, our proposed II framework is applied post-training at decision-time on the initial predictions of the default agent. We therefore "pre-train" our DreamerV3 agent using the default training settings described in \citet{hafner2023dreamerv3}, with the enabled latent disagreement exploration policy to obtain the necessary ensembles. The latter was already implemented as per \citet{sekar2020plan2expl} within \hyperlink{https://github.com/danijar/dreamerv3}{the official DreamerV3 re-implementation}. For evaluation at various stages of training, we saved a copy of the agent's parameters every 100K environment steps up to 1M, then every 500K steps after that. All of our agents make use of the large model architecture \citep{hafner2023dreamerv3}, for direct comparison between tasks.

\subsection{Evaluation}
For the agent using II, we applied $n=10$ II gradient steps after every environment step. The inference objective is calculated over $s=3$ sampled rollouts, of which the length is varied across experiments with values $\lambda \in \{1,3,8,16\}$. The added regularising term (Eq. \ref{eq_klReg}) is scaled such that its output is approximately of the same order of magnitude as the objective. 

For each pre-training volume, we varied the inference objective and rollout length $\lambda$.  We monitored the average inference objective values across all environment steps, for each gradient step. This allowed us to tune step-size $\alpha$ accordingly over a small number of episodes before running a full experiment. Inference objectives are averaged over the rollout length (see eq. \ref{eq_IG}-\ref{eq_entropyObj}) and number of sampled rollouts, to reduce the need for manual tuning of the step size $\alpha$ between experiments. Furthermore, we scale each objective to have similar orders of magnitude, as is done for the regularising term. Note that we use smaller step size values on larger pre-training volumes, as gradient steps that are too large seemed to have a tendency to \textit{increase} the objective loss during initial testing.

\textbf{Performance Metrics}\newline
Task performance and episode length are two common metrics to consider within RL environments. Although in most of our tasks longer episode length is directly linked to rewards, we make the assumption that, where appropriate, shorter episodes that achieve a similar or better score than longer counterparts are preferable as they can be seen as more efficient in completing the task.

Furthermore, as we are looking to improve the agent's state representations, we will also look at the impact on reconstruction metrics, comparing the observed $x_t$ with the decoder-reconstructed $\hat{x}_t$. To do so we will record the mean-squared error (MSE) between the two images, as well as two other common, and more robust metrics: the Peak-Signal-to-Noise-Ratio (PSNR) and the Structured Similarity Index (SSIM) \citep{ssim2004}. 

While the MSE has long been a standard full-reference quality metric, from which the PSNR is directly derived, it is very limited in judging the fidelity to the underlying structure of an image \citep{mse2009loveorleave}. The SSIM, on the other hand, is focused on the fidelity of the underlying structure and is less sensitive to non-structural distortions in the image (brightness, contrast, colour, etc..) (\citet{ssim2004}, \citet{mse2009loveorleave}).

As our agent must learn a compact representation of its sensory inputs to be used for action selection, we expect that the most important aspect of this representation for downstream task performance will be to accurately represent the information that is most relevant to successfully completing the task. In the case of navigating a visual environment, we assume that structural fidelity will be most relevant to the agent's task performance. We therefore take SSIM to measure any potential changes to this "actionable" information within our latent state, and record the MSE and PSNR as a more general measure of total information recall. Note that we seek to \textit{minimise} MSE, and \textit{maximise} PSNR and SSIM.

Reconstruction metrics are recorded at every environment step on the agents' initial predictions, before applying the II updates, as well as on the updated predictions. This is done to measure any potential immediate impact on the latent state. As each episode has a random initialisation, we record our performance metrics over a minimum of $100$ episodes for each experiment. This allows us to make more confident comparisons with our baselines.

\subsection{Computing Environments}\label{app_computingEnvironments}
Our agents were pre-trained using a V100 GPU in order to accelerate training and replicate the conditions described in the original DreamerV3 paper \citep{hafner2023dreamerv3}. This hardware was accessed using the hosted Google Colaboratory platform \footnote{https://colab.google/, as of January 2024}. Debugging and initial testing was completed using simple CPU run-times, and evaluation was done using either a V100 or T4 GPU, depending on the complexity resulting from our hyper-parameters: very deep rollouts with many gradient steps would take much longer to run than a shallower approach on the slower, but cheaper, T4 GPU. 

\newpage
\section{Environments \& Tasks}\label{app_env_tasks}
\FloatBarrier
\subsection{DMLab}

\subsubsection{Collect Good Objects}\label{app_collectGood}

\begin{figure}[tbh!]
    \centering
    \includegraphics{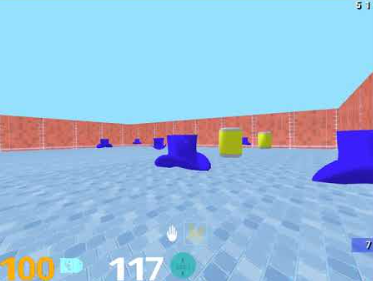}
    \caption{DMLab's \textit{Collect Good Objects task}. \href{https://github.com/google-deepmind/lab/blob/master/game_scripts/levels/contributed/dmlab30/README.md}{Source}.}
    \label{fig_task_collectgood}
\end{figure}

The task "Collect Good Objects", as originally described in \citet{higgins2018darla}, consists in a seek-avoid object collection setup, where "good" objects must be collected for a positive reward ($+1$), while "bad" objects incur a negative reward ($-1$). The collection of 10 objects, whether good or bad, terminates the episode. If 10 objects aren't collected, the episode terminates after 900 steps. 
There are two room types (pink and green by wall colour), and four object types (balloon, cake, can, and hat). The agent is trained with a subset of all possible room-objects combinations, and is presented with \textit{unseen} environments at test time to measure its ability to transfer knowledge to unfamiliar environments. These room-object combinations are generated at random when creating the environment for evaluation. This is beneficial for our own experiments as system 1 should struggle to infer in new environments such as this, while system 2 should be beneficial in supporting the agent's ability to generalise to this unseen setting.

\FloatBarrier
\subsubsection{NatLab Fixed Large Map}\label{app_natlab}

\begin{figure}[tbh!]
    \centering
    \includegraphics[width=0.6\textwidth]{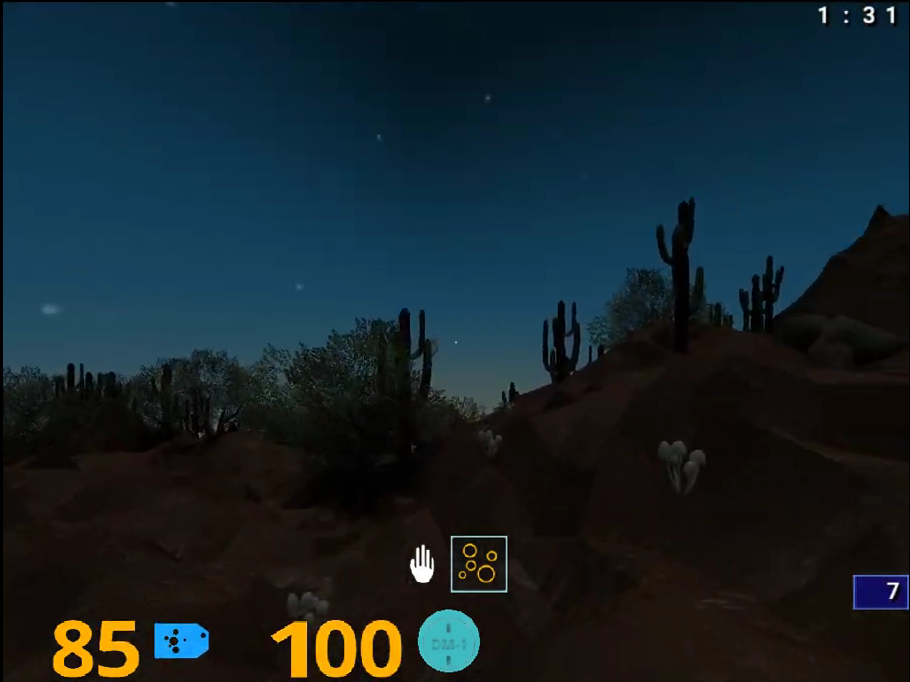}
    \caption[NatLab]{DMLab's \textit{NatLab Fixed Large Map task}. \href{https://www.youtube.com/watch?v=ucJEnnn5iC8}{Source}.}
    \label{fig_task_natlab}
\end{figure}

The agent must collect as many mushrooms as possible within a complex naturalistic terrain environment to maximise score. The mushrooms do not regrow, as opposed to other variations of this task. The large map layout is fixed regardless of initialisation. The time of day is randomised (day, dawn, night), which varies the lighting and colours observed. Furthermore, the spawn location is picked randomly from a set of potential spawn locations at initialisation. The episode terminates after 1800 environment steps. We note that there are no separate training and testing regimes, with all episodes drawn from the same distribution.

\FloatBarrier
\subsection{Atari}
\subsubsection{Alien}\label{app_atari_alien}

\begin{figure}[bt!]
    \centering
    \includegraphics[height=0.3\textheight]{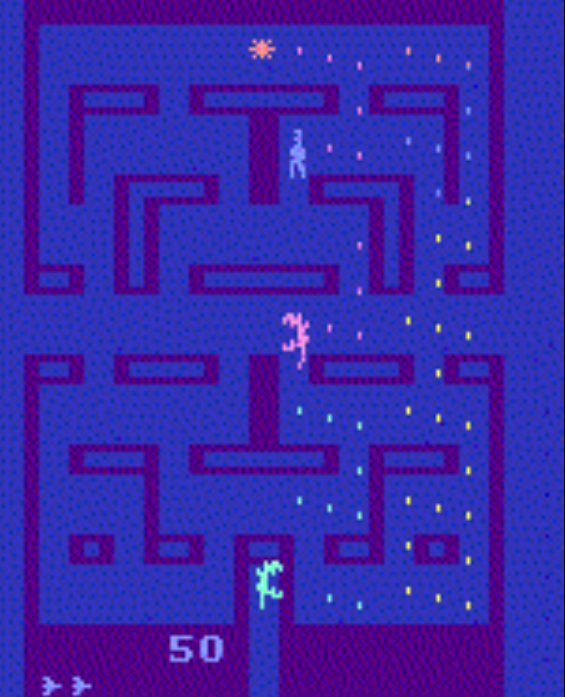}
    \caption{Atari's \textit{Alien} task. \href{https://www.gymlibrary.dev/environments/atari/alien/}{Source}.}
    \label{fig_task_atari}
\end{figure}

The agent acts in a 2D, "maze-like" map with three aliens. The goal is to destroy their eggs (white pixels, fig. \ref{fig_task_atari}) simply by running over them. These are evenly scattered all over the map. Destroying all eggs on the map leads to a new, more difficult map. The agent must simultaneously avoid the aliens, which are chasing it. The agent possesses a flamethrower that can help turn away the aliens. The agent starts with two lives, and loses a life when an alien catches it. Losing both lives terminates the episode.

\FloatBarrier
\subsection{Miniworld}
\subsubsection{Robot Navigation}\label{app_roboNav}

We construct a custom environment built off the well-known Miniworld library \citep{gym_miniworld}, specifically designed to simulate vision-based robotic navigation to a randomly positioned static target. Unlike standard Miniworld environments, our custom environment incorporates both static and dynamically moving obstacles which the agent must avoid to prevent incurring negative rewards. Furthermore, we incorporate a checkpoint based system for positive rewards, such that the agent receives one third of the max reward (modified by a time-based decay factor) when one third of the initial distance to the target is crossed, another one third at two thirds the distance, and a final one third when it makes contact with the target. This reward structure was chosen because the sparsity of a single positive reward was found to be too difficult for even the largest of Dreamer models both with and without iterative inference. 

For initial distance to target, $D$, current distance to target $d_t$, and max episode length $T$, the reward structure may be described as follows:
\begin{align}
    \mathcal{R}_t &= 
    \begin{cases}
    +\frac{1}{3}\cdot(1 - 0.2 \cdot (t / T))  & \text{if } d_t \in {\frac{2}{3}D, \frac{1}{3}D, 0} \\
    -0.05 & \text{If touching a static obstacle} \\
    -0.1 & \text{If touching a dynamic obstacle}
    \end{cases}
\end{align}

\begin{figure}[!htb]
    \centering
    \includegraphics[width=0.8\linewidth, keepaspectratio]{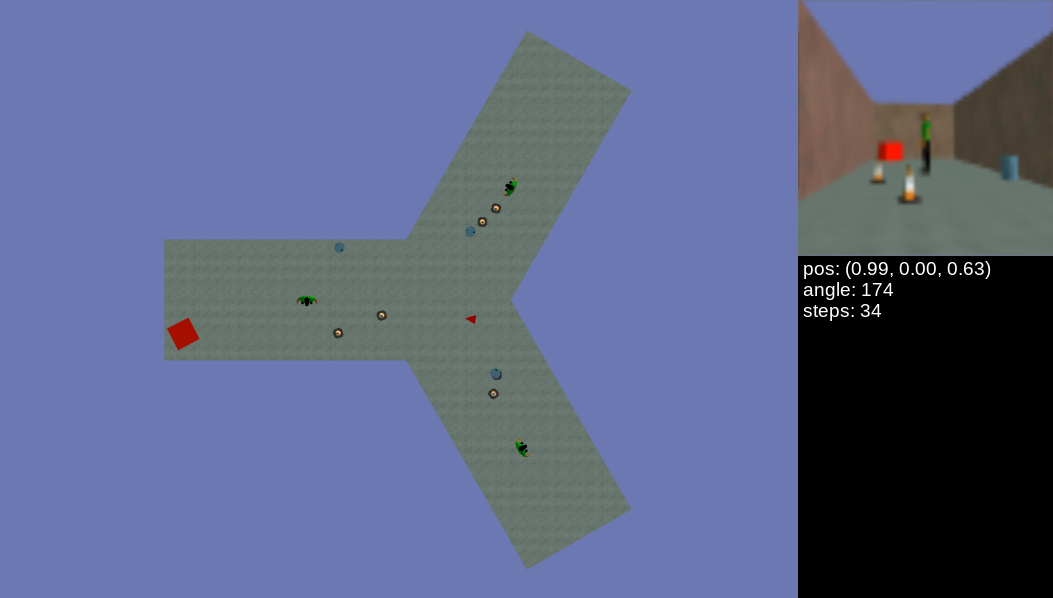}
    \caption[Robot Navigation Environment]{(Left) Top down view of our robot navigation environment. (Right top) RGB input as provided to the agent. (Right bottom) Agent position, angle, and number of environment steps; not provided as observations to agent. }
    \label{fig_robotNavEnvironment}
\end{figure}

All three possible positive rewards are only granted once per episode, with each episode having a maximum of 400 steps before termination. The environment structure takes the form of a Y Maze, with domain randomisation over the target position, static obstacle positions, number of static obstacles in each room (from 0 to 3), static obstacle mesh (cone or barrel), sky colour, light position and light colour. Target positions were randomised to the posterior third of one of the three hallways chosen at random each episode, moving obstacles were randomised to the middle third, and static obstacles were randomised to the anterior third of the hallways. A top-down view and example observation input can be seen in Figure \ref{fig_robotNavEnvironment}. This structure ensured that the agent had to first identify which hallway contained the red box, and avoid both static and dynamic obstacles to successfully reach it. 

\newpage
\section{Supplementary Figures}\label{app_fullresults}
\subsection{DMLab: Collect Good Objects}

\begin{figure}[!htb]
    \centering
    \includegraphics[width=\linewidth, height=0.5\textheight, keepaspectratio]{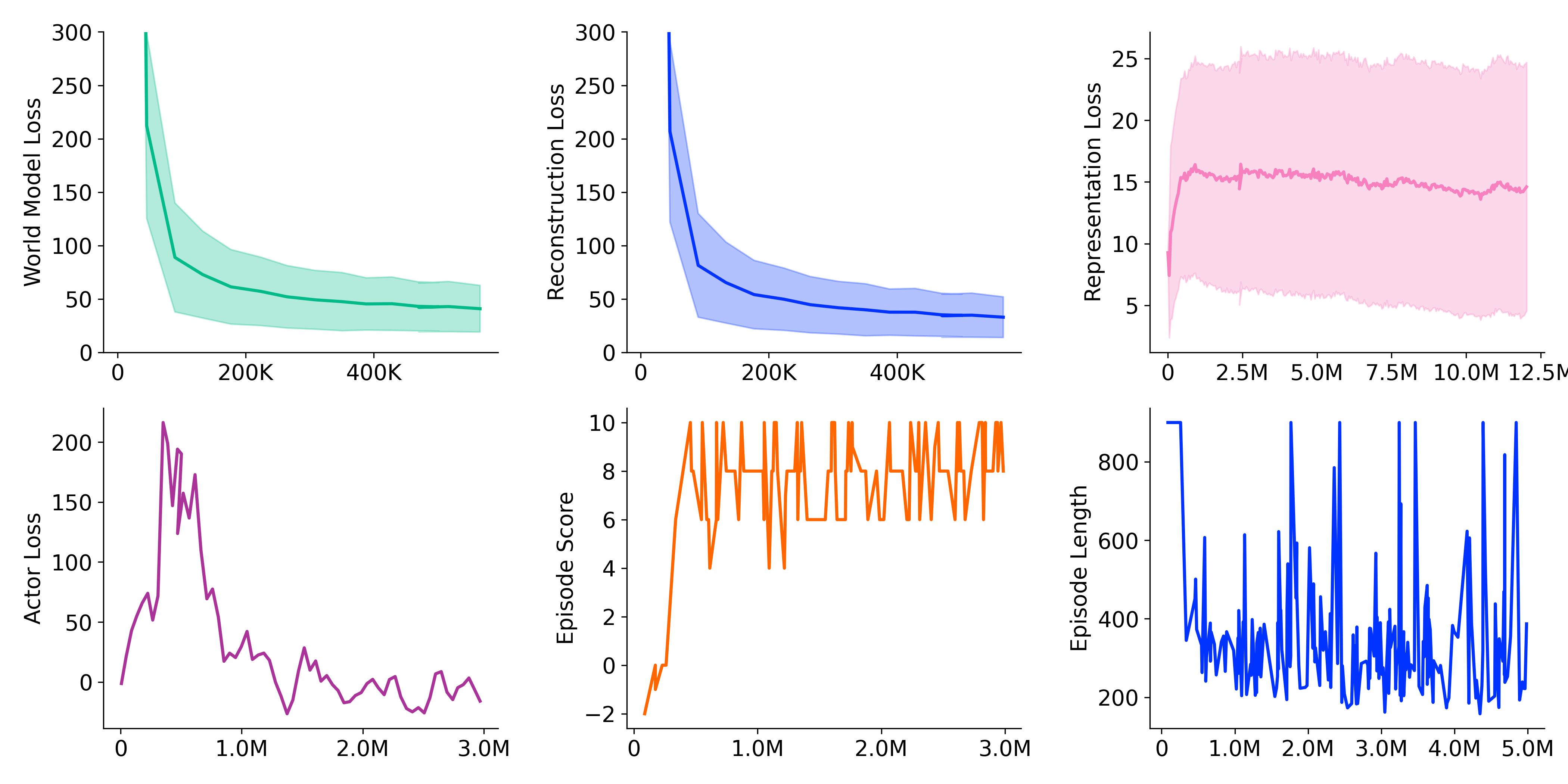}
    \caption[Pre-training Metrics over Environment Steps]{Losses and metrics over environment steps during pre-training of DreamerV3 on the DMLab \textit{Collect Good Objects} task. We stopped training at 30M environment steps, but have cropped the above plots for a better visual of where key changes in values occur.}
    \label{fig_pretrainingLosses_dmlab}
\end{figure}

\begin{figure}[!htb]
    \centering
    \includegraphics[width=\linewidth, height=0.5\textheight, keepaspectratio]{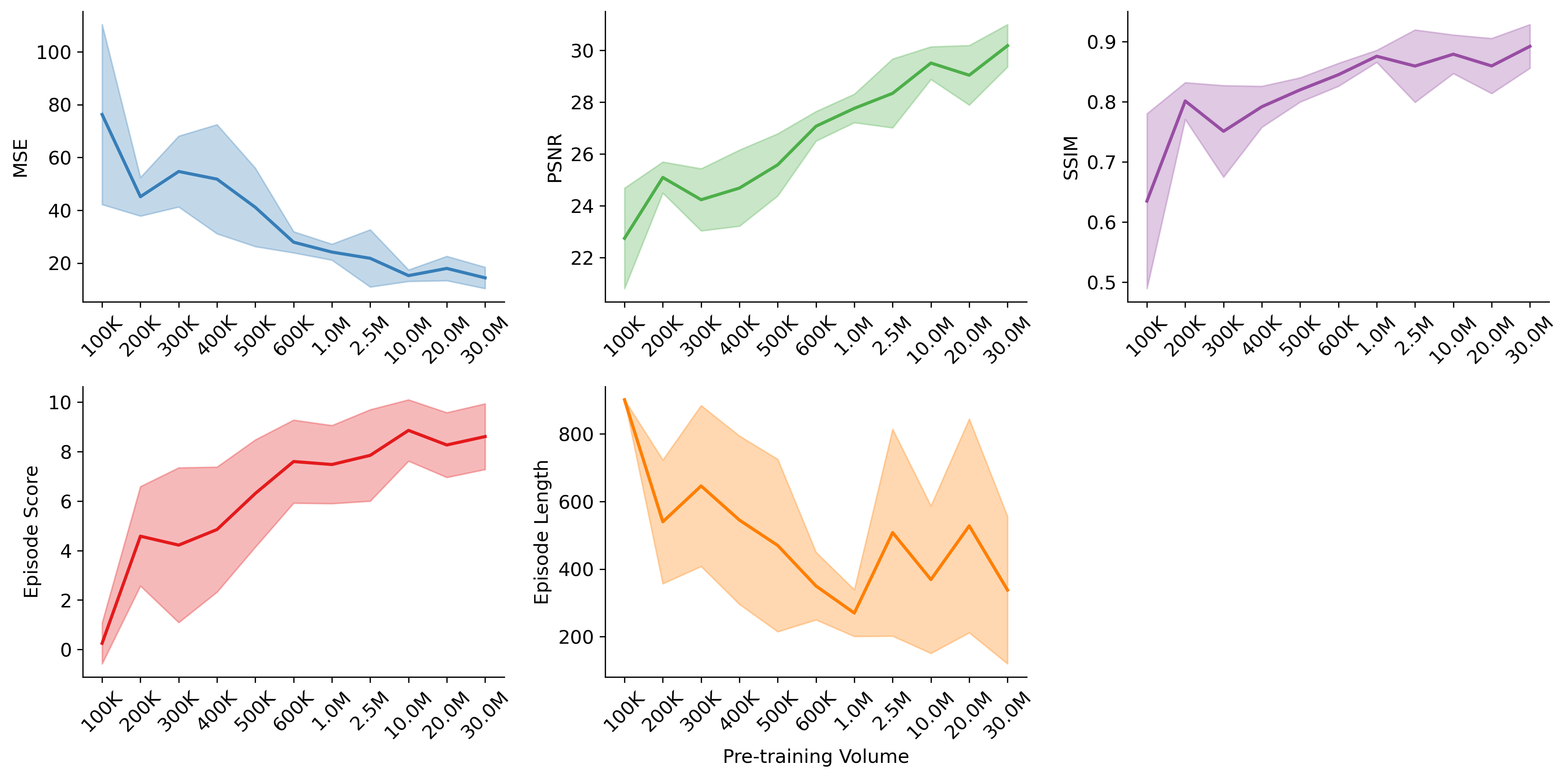}
    \caption[Comparing Baseline Performance v. Pre-Training Volume]{Average performance metrics for the default DreamerV3 agent baseline on the DMLab \textit{Collect Good Objects} task for various pre-training volumes. Metrics are averaged over $100$ episodes for each volume, and standard deviation is measured across episodes.}
    \label{fig_baselinePerformance_dmlab}
\end{figure}

\begin{figure}[!htb]
    \centering
    \includegraphics[width=\linewidth, height=0.5\textheight, keepaspectratio]{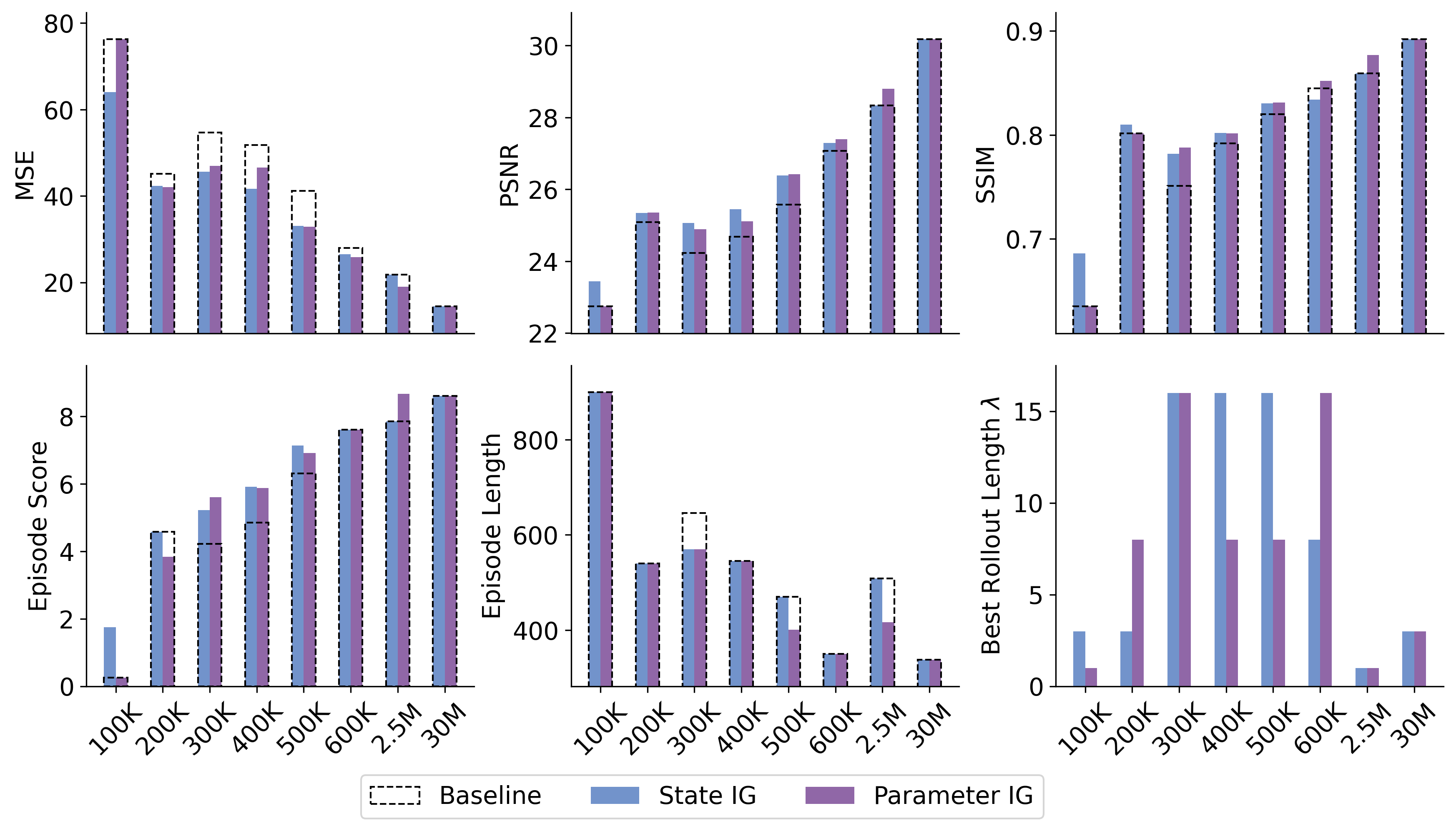}
    \caption[Inference Objective Performance Comparison v. Pre-Training Volume]{Comparison of performance metrics for the default DreamerV3 agent (Baseline), versus iterative inference with two inference objectives, on the DMLab \textit{NatLab Fixed Large Map} task. For each objective, we take the best, \textit{significant} improvement compared to the baseline (ie. a p-value below 5\%), across all metrics and rollout lengths. If the best performance shows no significant improvement or decrease in performance, we assume the same value as the baseline. "Best Rollout Length $\lambda$" shows which value of $\lambda$ yielded the best performance for each objective, and hence, which experiment is shown in the other plots.}
    \label{fig_full_dmlab}
\end{figure}

\FloatBarrier
\clearpage
\newpage
\subsection{DMLab: NatLab Fixed Large Map}

\begin{figure}[!htb]
    \centering
    \includegraphics[width=\linewidth, height=0.5\textheight, keepaspectratio]{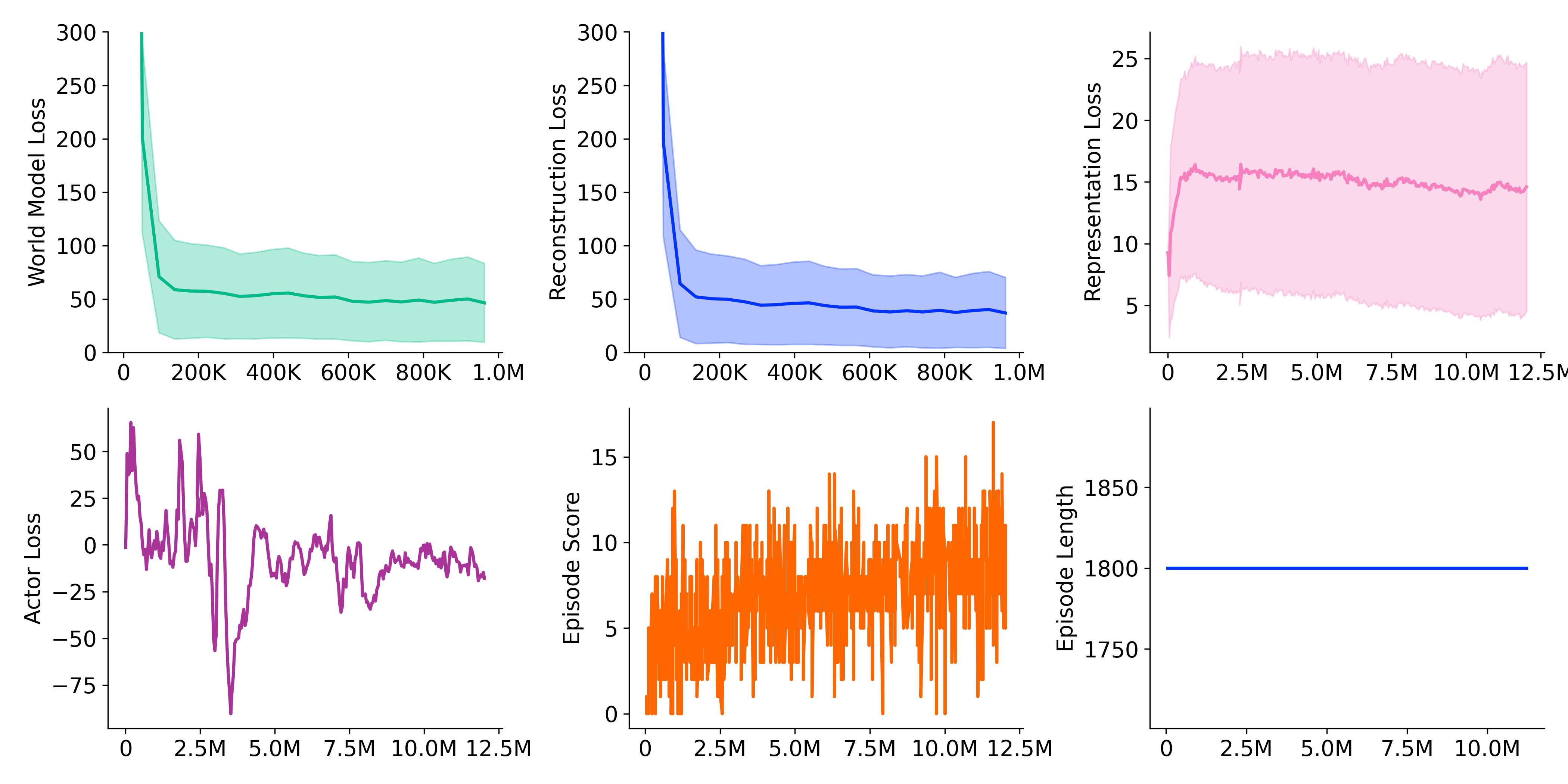}
    \caption[Pre-training Metrics over Environment Steps]{Losses and metrics over environment steps during pre-training of DreamerV3 on the DMLab \textit{NatLab Fixed Large Map} task. We stopped training at 12M environment steps, but have cropped the above plots for a better visual of where key changes in values occur.}
    \label{fig_pretrainingLosses_natlab}
\end{figure}

\begin{figure}[!htb]
    \centering
    \includegraphics[width=\linewidth, height=0.5\textheight, keepaspectratio]{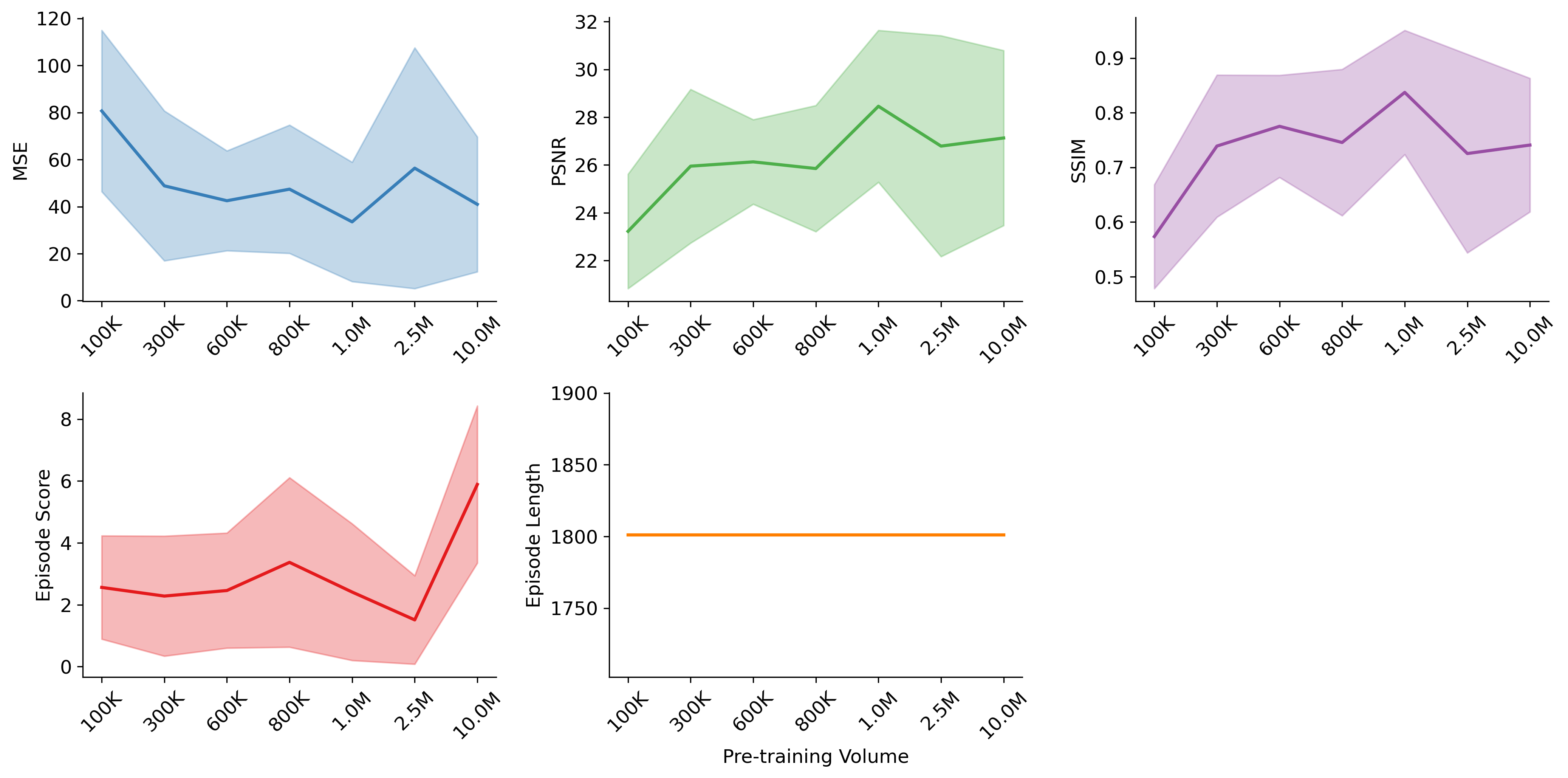}
    \caption[Comparing Baseline Performance v. Pre-Training Volume]{Average performance metrics for the default DreamerV3 agent baseline on the DMLab \textit{NatLab Fixed Large Map} task for various pre-training volumes. Metrics are averaged over $100$ episodes for each volume, and standard deviation is measured across episodes.}
    \label{fig_baselinePerformance_natlab}
\end{figure}

\begin{figure}[!htb]
    \centering
    \includegraphics[width=\linewidth, height=0.5\textheight, keepaspectratio]{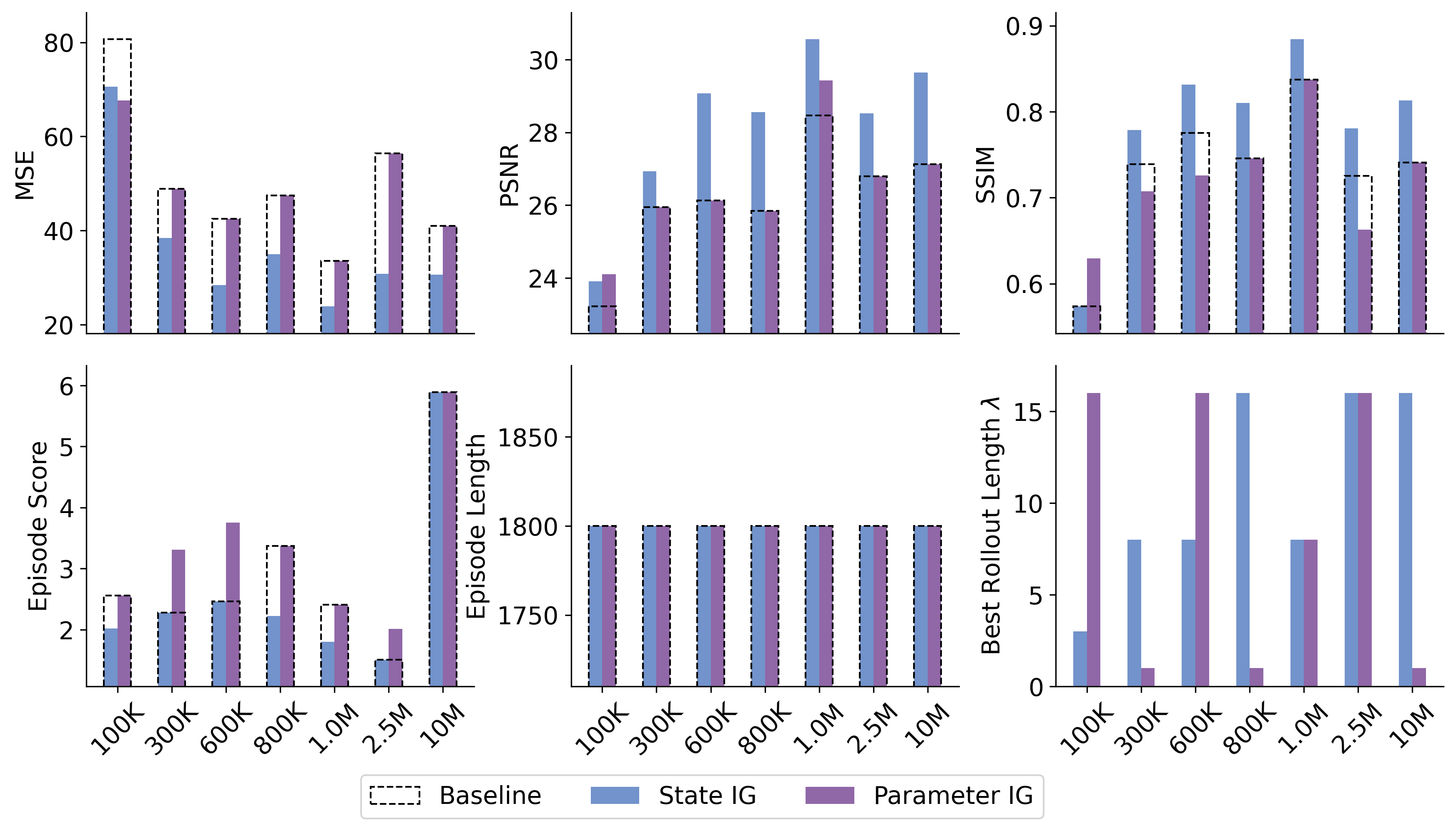}
    \caption[Inference Objective Performance Comparison v. Pre-Training Volume]{Comparison of performance metrics for the default DreamerV3 agent (Baseline), versus iterative inference with two inference objectives, on the DMLab \textit{NatLab Fixed Large Map} task. For each objective, we take the best, \textit{significant} improvement compared to the baseline (ie. a p-value below 5\%), across all metrics and rollout lengths. If the best performance shows no significant improvement or decrease in performance, we assume the same value as the baseline. "Best Rollout Length $\lambda$" shows which value of $\lambda$ yielded the best performance for each objective, and hence, which experiment is shown in the other plots.}
    \label{fig_full_natlab}
\end{figure}

\FloatBarrier
\clearpage
\newpage
\subsection{Atari: Alien}

\begin{figure}[!htb]
    \centering
    \includegraphics[width=\linewidth, height=0.5\textheight, keepaspectratio]{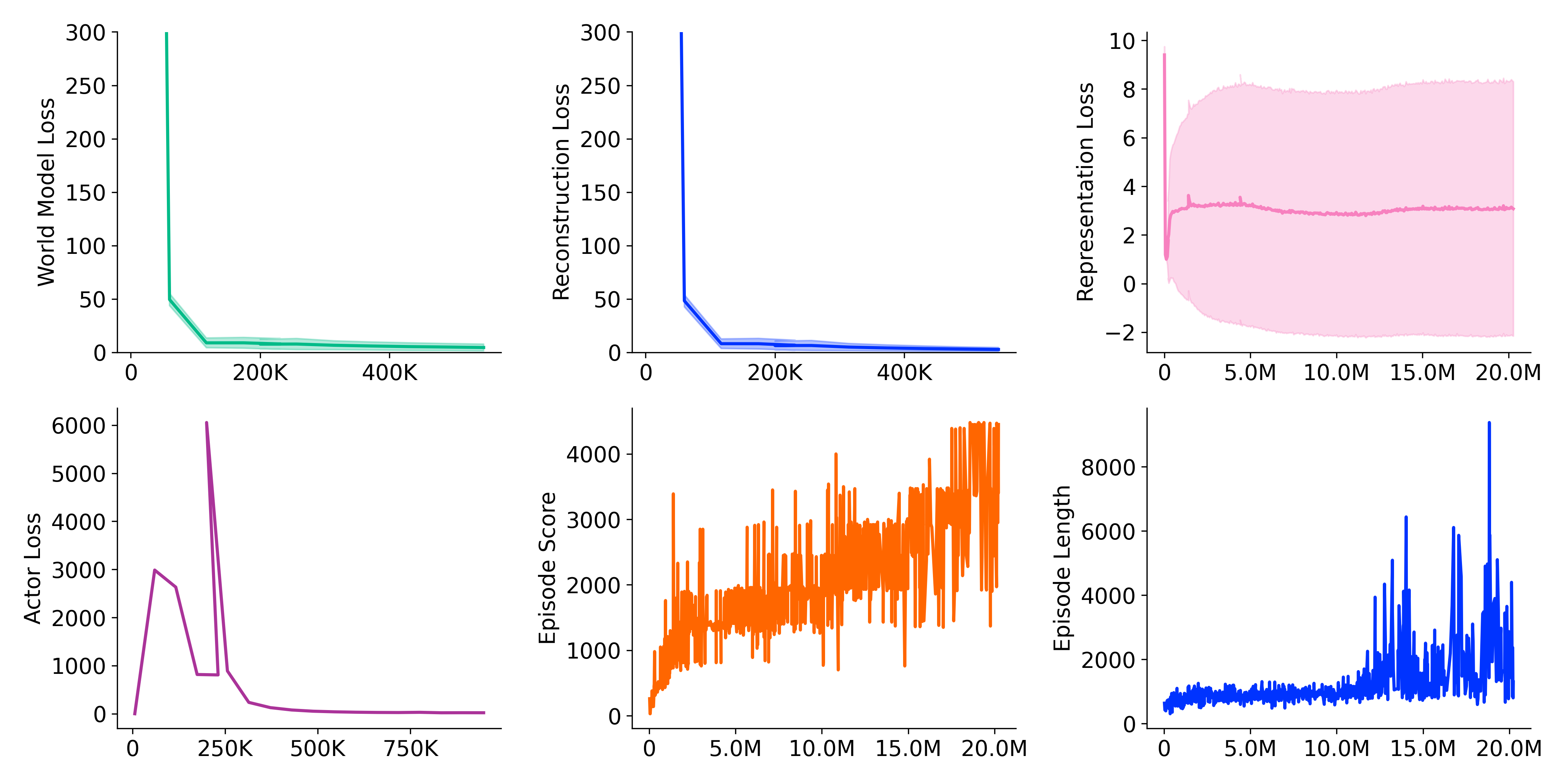}
    \caption[Pre-training Metrics over Environment Steps]{Losses and metrics over environment steps during pre-training of DreamerV3 on the Atari \textit{Alien} task. We stopped training at 12M environment steps, but have cropped the above plots for a better visual of where key changes in values occur.}
    \label{fig_pretrainingLosses_alien}
\end{figure}

\begin{figure}[!htb]
    \centering
    \includegraphics[width=\linewidth, height=0.5\textheight, keepaspectratio]{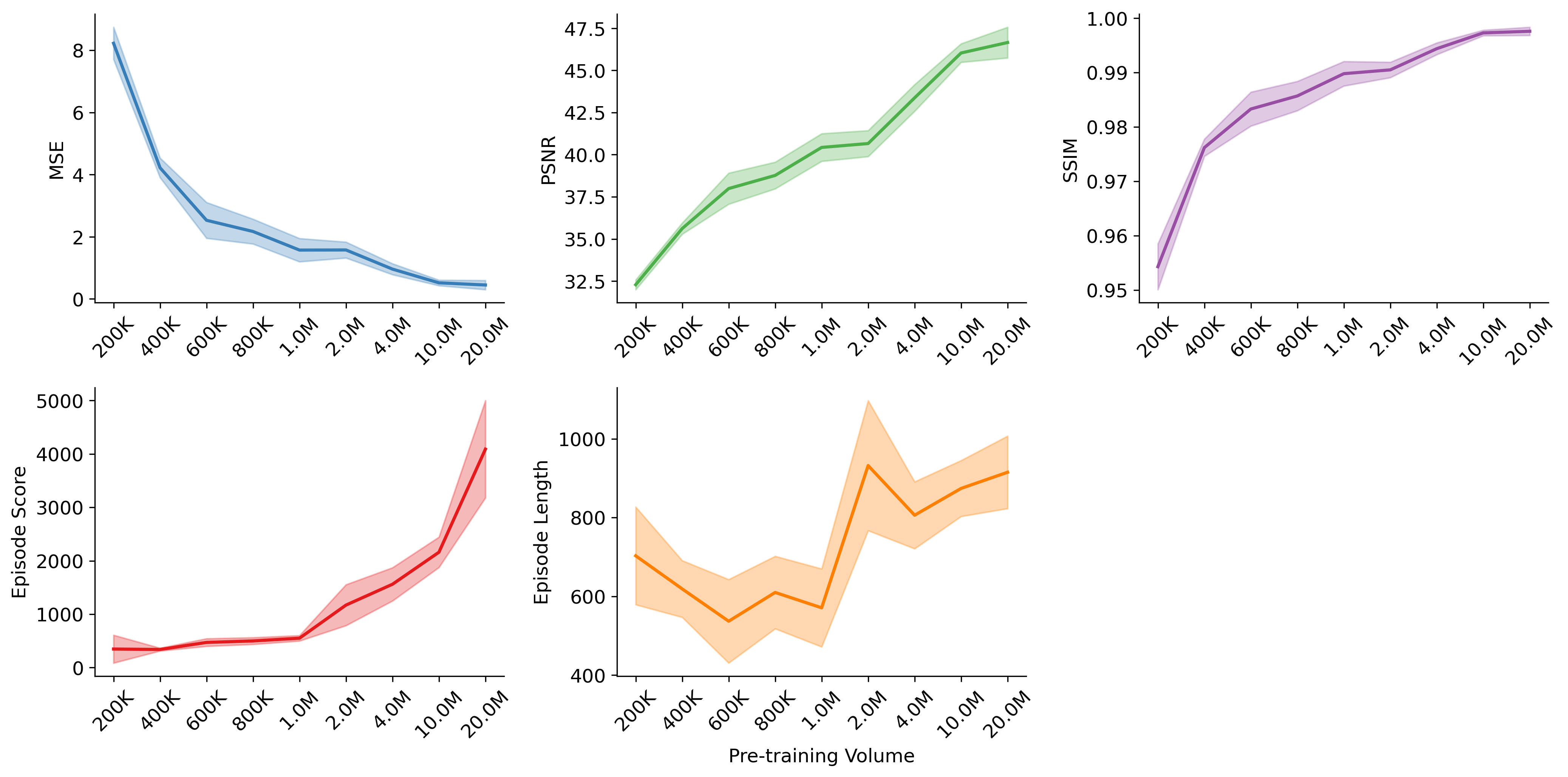}
    \caption[Comparing Baseline Performance v. Pre-Training Volume]{Average performance metrics for the default DreamerV3 agent baseline on the Atari \textit{Alien} task for various pre-training volumes. Metrics are averaged over $100$ episodes for each volume, and standard deviation is measured across episodes.}
    \label{fig_baselinePerformance_atari}
\end{figure}

\begin{figure}[!htb]
    \centering
    \includegraphics[width=\linewidth, height=0.5\textheight, keepaspectratio]{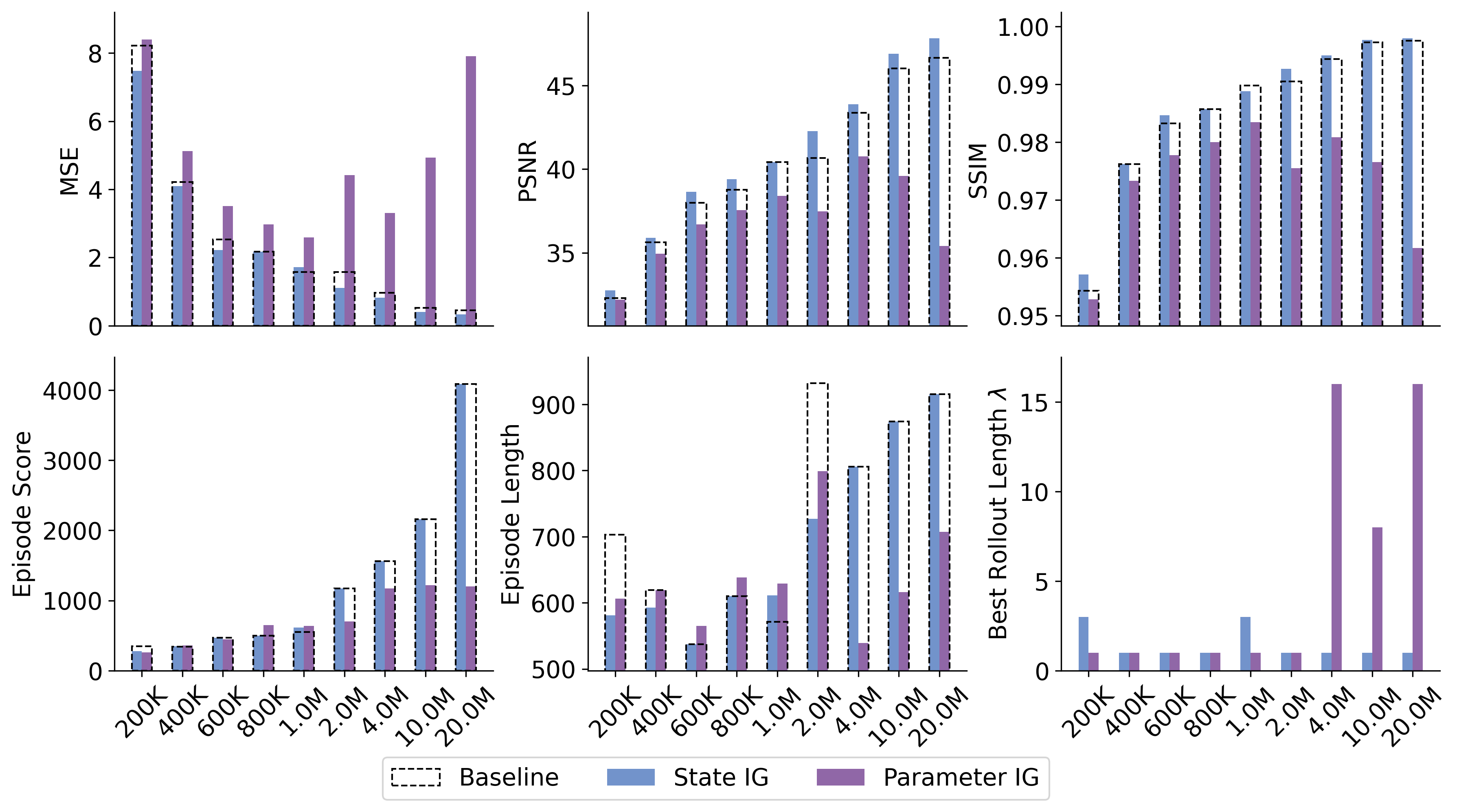}
    \caption[Inference Objective Performance Comparison v. Pre-Training Volume]{Comparison of performance metrics for the default DreamerV3 agent (Baseline), versus iterative inference with two inference objectives, on the Atari \textit{Alien} task. For each objective, we take the best, \textit{significant} improvement compared to the baseline (ie. a p-value below 5\%), across all metrics and rollout lengths. If the best performance shows no significant improvement or decrease in performance, we assume the same value as the baseline. "Best Rollout Length $\lambda$" shows which value of $\lambda$ yielded the best performance for each objective, and hence, which experiment is shown in the other plots.}
    \label{fig_full_alien}
\end{figure}

\FloatBarrier
\subsection{Miniworld: Robot Navigation (our task)}

\begin{figure}
    \centering
    \includegraphics[width=0.8\linewidth, height=0.4\textheight, keepaspectratio]{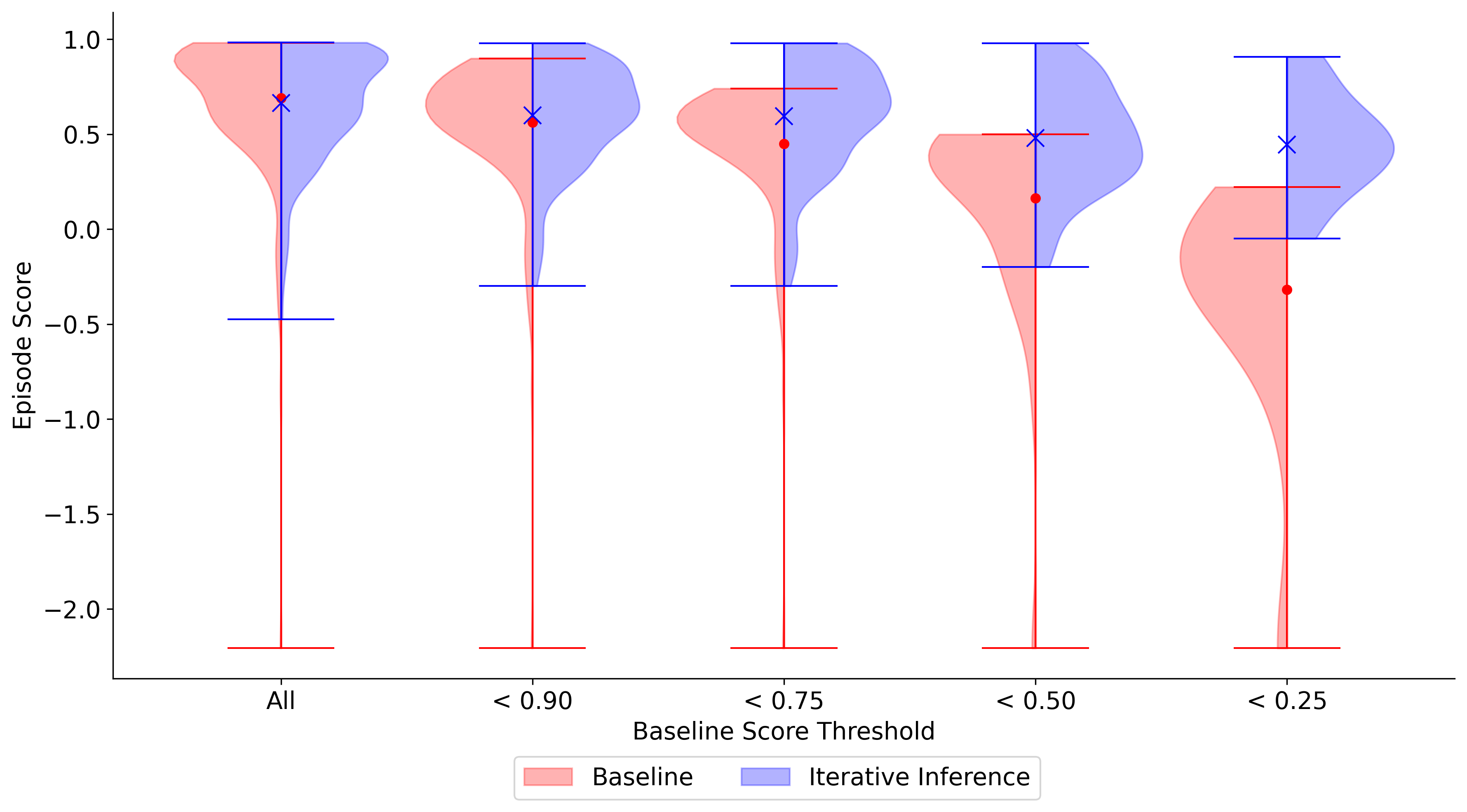}
    \caption{Miniworld task performance after 1M pre-training steps with State IG ($\lambda = 16$), in episodes with the same initialisation. "Baseline Score Threshold" shows the baseline agent score threshold with which we selected episodes. We therefore compare II and baseline only on episodes where baseline scored lower than the threshold. The red dot and blue cross show the average scores for baseline and II respectively. This figure compares a total of 250 episodes.}
    \label{fig_roboNav_matched_lessthan}
\end{figure}

\begin{figure}
    \centering
    \includegraphics[width=0.8\linewidth, height=0.4\textheight, keepaspectratio]{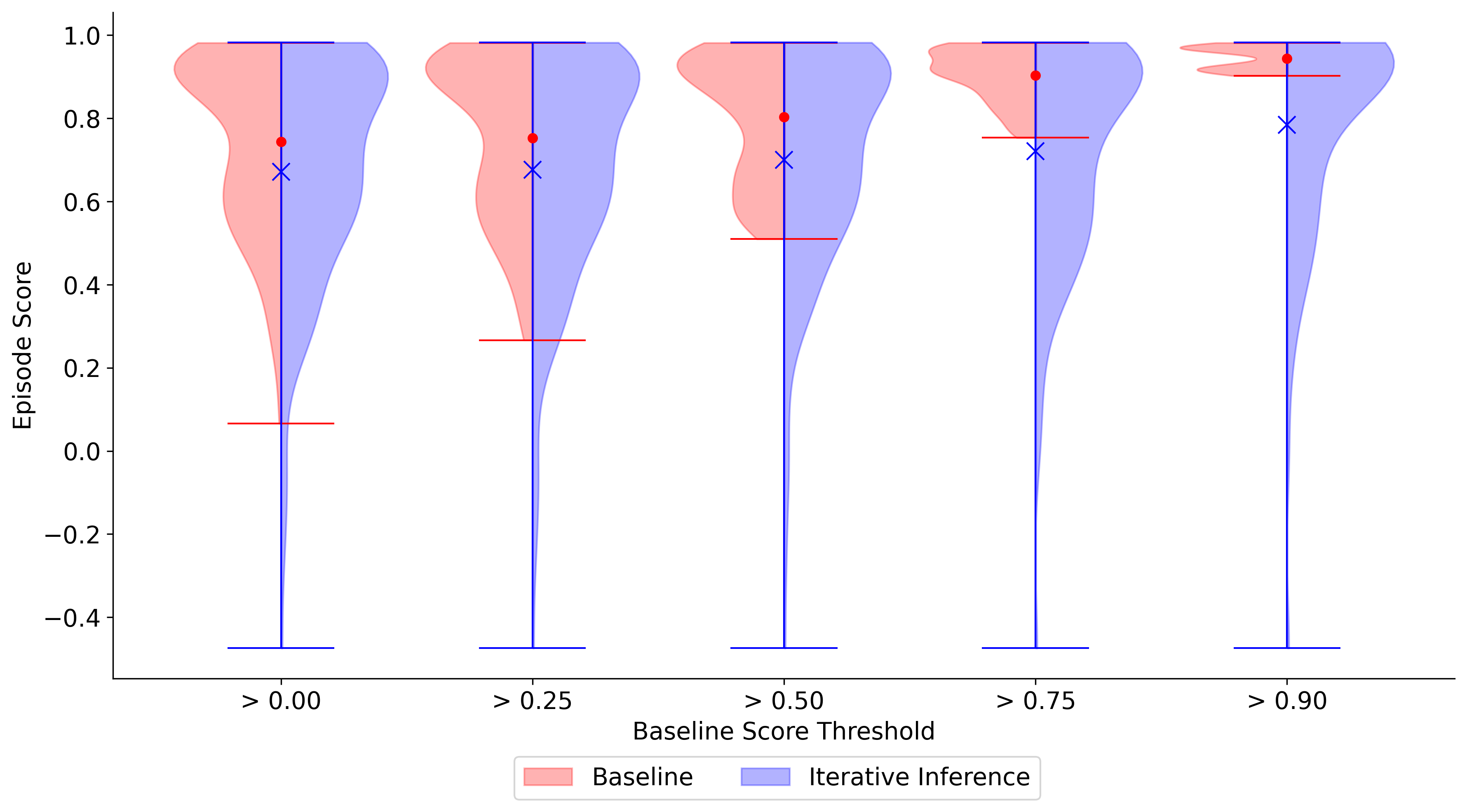}
    \caption{Miniworld task performance after 1M pre-training steps with State IG ($\lambda = 16$), in episodes with the same initialisation. "Baseline Score Threshold" shows the baseline agent score threshold with which we selected episodes. We therefore compare II and baseline only on episodes where baseline scored greater than the threshold. The red dot and blue cross show the average scores for baseline and II respectively. This figure compares a total of 250 episodes.}
    \label{fig_roboNav_matched_greaterthan}
\end{figure}

\begin{figure}
    \centering
    \includegraphics[width=0.8\linewidth, height=0.4\textheight, keepaspectratio]{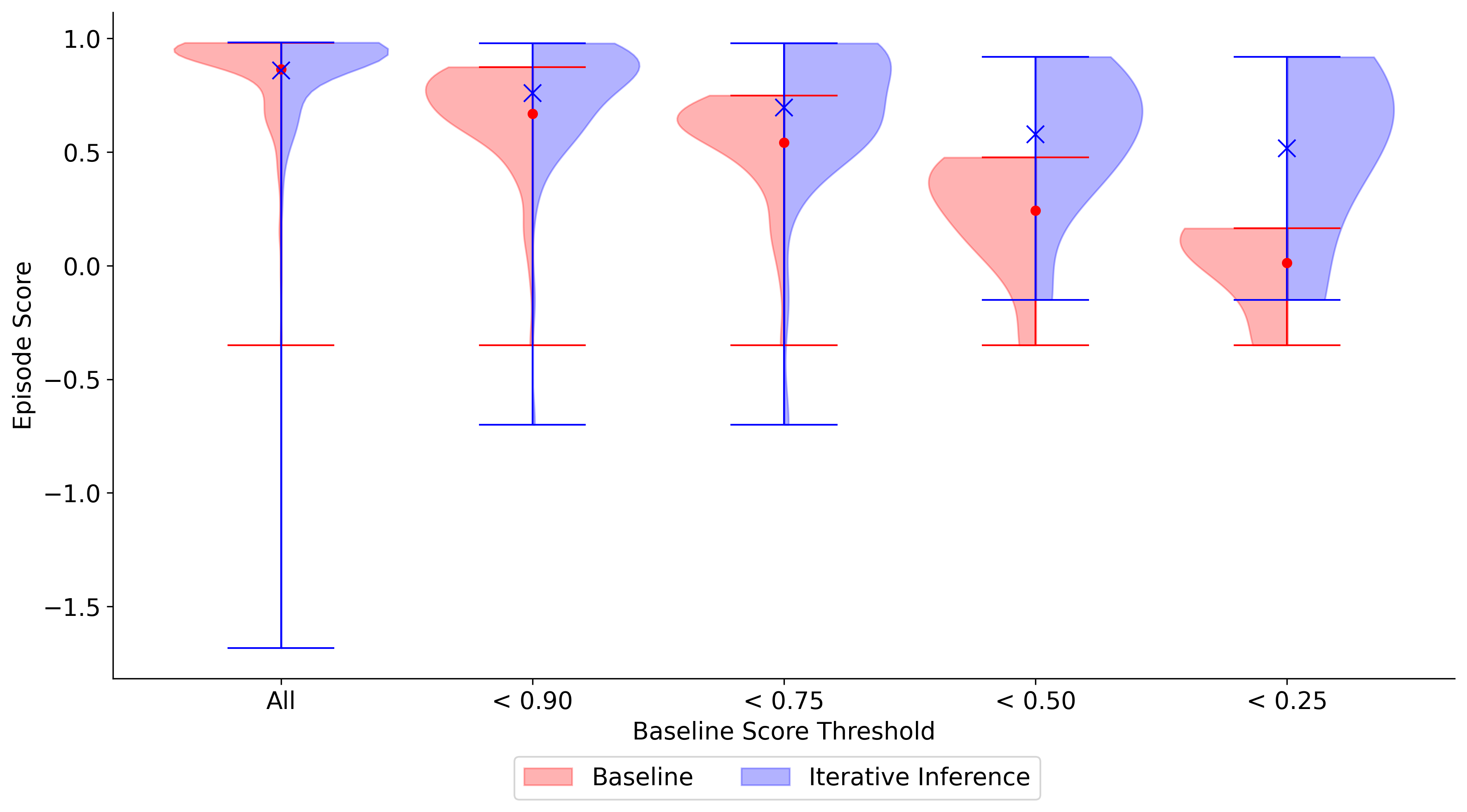}
    \caption{Miniworld task performance after 2.5M pre-training steps with Parameter IG ($\lambda = 16$), in episodes with the same initialisation. "Baseline Score Threshold" shows the baseline agent score threshold with which we selected episodes. We therefore compare II and baseline only on episodes where baseline scored lower than the threshold. The red dot and blue cross show the average scores for baseline and II respectively. This figure compares a total of 250 episodes.}
    \label{fig_roboNav_matched_lessthan_2.5M}
\end{figure}

\begin{figure}
    \centering
    \includegraphics[width=\linewidth]{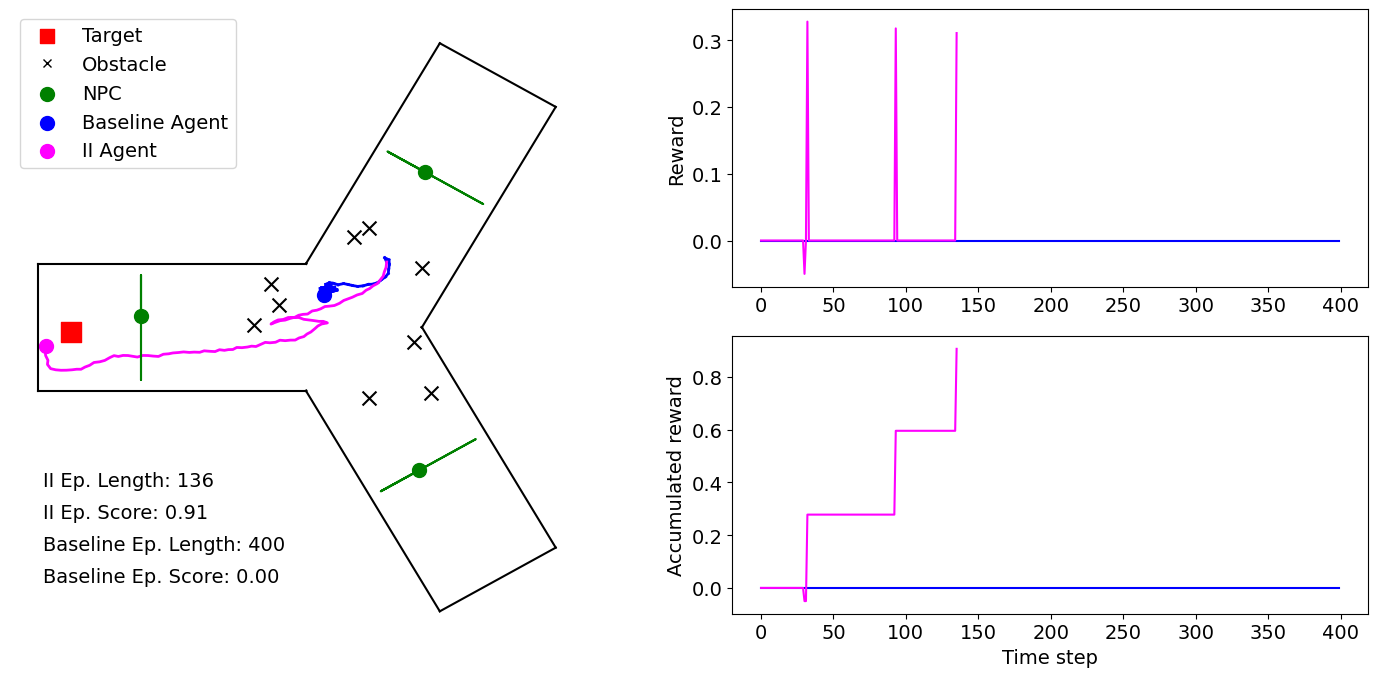}
    \caption{Comparing the behaviour of II vs the baseline agent in a single episode with the same initialisation and layout. 1M pre-training steps with the State IG objective and $\lambda = 16$.}
    \label{fig_robonav_traj_1}
\end{figure}

\begin{figure}
    \centering
    \includegraphics[width=\linewidth]{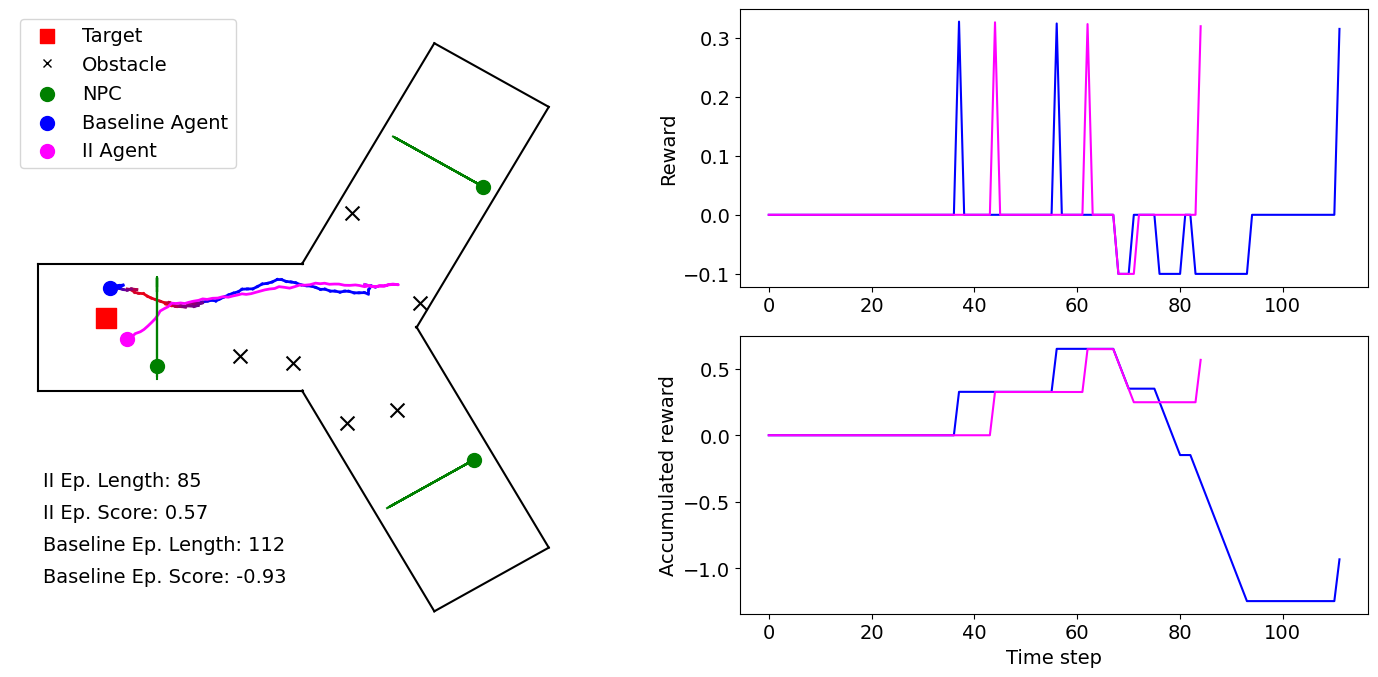}
    \caption{Comparing the behaviour of II vs the baseline agent in a single episode with the same initialisation and layout. 1M pre-training steps with the State IG objective and $\lambda = 16$.}
    \label{fig_robonav_traj_2}
\end{figure}

\begin{figure}
    \centering
    \includegraphics[width=\linewidth]{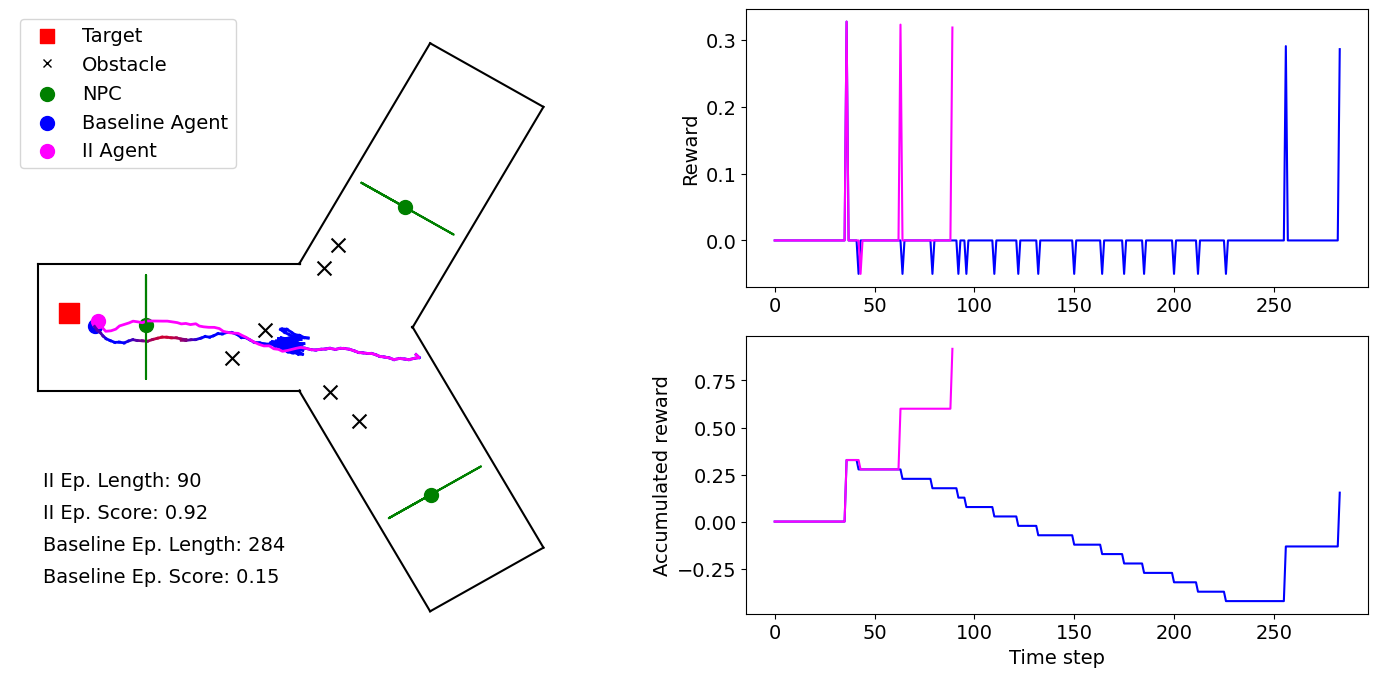}
    \caption{Comparing the behaviour of II vs the baseline agent in a single episode with the same initialisation and layout. 2.5M pre-training steps with the Parameter IG objective and $\lambda = 16$.}
    \label{fig_robonav_traj_3}
\end{figure}

\begin{figure}
    \centering
    \includegraphics[width=\linewidth]{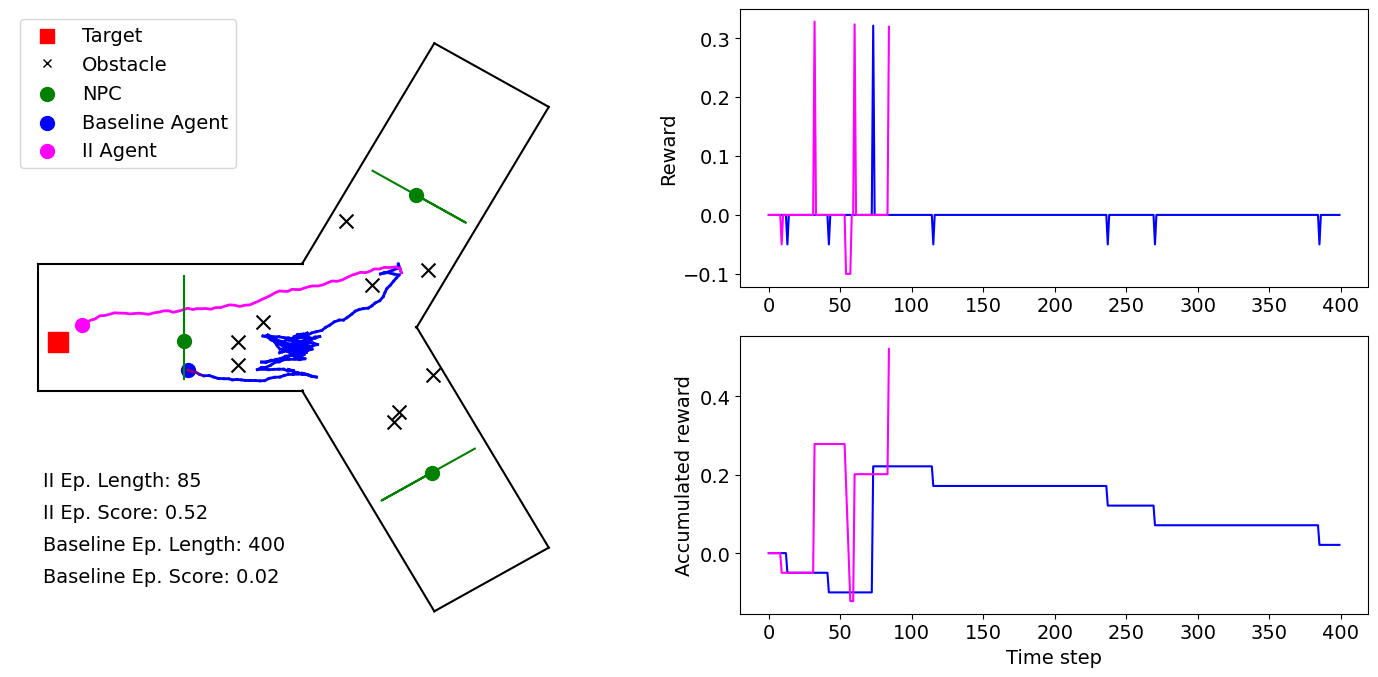}
    \caption{Comparing the behaviour of II vs the baseline agent in a single episode with the same initialisation and layout. 2.5M pre-training steps with the State IG objective and $\lambda = 16$.}
    \label{fig_robonav_traj_4}
\end{figure}

\end{document}